\documentclass[twoside]{article}

\usepackage[accepted]{aistats2026}

\usepackage{natbib}

\setcitestyle{numbers,sort,round}
\setcitestyle{authoryear,open={(},close={)}, aysep={,}} 
\usepackage{cleveref}
\usepackage{xcolor}
\usepackage{algorithm}
\usepackage{algpseudocode}
\usepackage{booktabs}       
\usepackage{graphicx} 
\usepackage{caption}
\usepackage{booktabs}  
\usepackage{url}
\usepackage{rotating}
\usepackage{nicefrac}       
\usepackage{microtype}
\usepackage{bm}
\usepackage{amsmath}
\usepackage{amsthm}
\theoremstyle{definition}
\theoremstyle{plain}
\newtheorem{theorem}{Theorem}[section]   
\newtheorem{lemma}[theorem]{Lemma}       

\usepackage{pdfpages}
\usepackage{enumitem}

\usepackage{subcaption}
\usepackage[normalem]{ulem}

\useunder{\uline}{\ul}{}

\usepackage{amssymb}
\usepackage{pifont}
\DeclareMathOperator*{\argmax}{arg\,max}
\DeclareMathOperator{\conv}{conv}

\newcommand{\eg}{e.\,g., }
\newcommand{\ie}{i.\,e., }
\theoremstyle{plain}

\begin{document}

%
\runningtitle{Archetypal Graph Generative Models}

%

\runningauthor{Nakis, Kosma, Promponas, Chatzianastasis, Nikolentzos}

\twocolumn[

\aistatstitle{Archetypal Graph Generative Models: Explainable and Identifiable Communities via Anchor-Dominant Convex Hulls}

\aistatsauthor{ Nikolaos Nakis$^\dagger$ \And Chrysoula Kosma$^\ddagger$ \And Panagiotis Promponas$^\dagger$}
\aistatsauthor{ Michail Chatzianastasis$^\clubsuit$ \And Giannis Nikolentzos$^\spadesuit$}

\aistatsaddress{ $^\dagger$Yale Institute for Network Science, Yale University, USA \\ $^\ddagger$Université Paris Saclay, Université Paris Cité, ENS Paris Saclay, CNRS, SSA, INSERM, Centre Borelli, France \\ $^\clubsuit$École Polytechnique, Paris, France
 \\ $^\spadesuit$Department of Informatics and Telecommunications, University of Peloponnese, Greece} ]

\begin{abstract}
Representation learning has been essential for graph machine learning
tasks such as link prediction, community detection, and network visualization. Despite recent advances in achieving high performance on these downstream tasks, little progress has been made toward self-explainable models. Understanding the patterns behind predictions is equally important, motivating recent interest in explainable machine learning. In this paper, we present \textsc{GraphHull}, an explainable generative model that represents networks using two levels of convex hulls. At the global level, the vertices of a convex hull are treated as \emph{archetypes}, each corresponding to a pure community in the network. At the local level, each community is refined by a \emph{prototypical hull} whose vertices act as representative profiles, capturing community-specific variation. This two-level construction yields clear multi-scale explanations: a node’s position relative to global archetypes and its local prototypes directly accounts for its edges. The geometry is well-behaved by design, while local hulls are kept disjoint by construction. To further encourage diversity and stability, we place principled priors, including determinantal point processes, and fit the model under MAP estimation with scalable subsampling. Experiments on real networks demonstrate the ability of \textsc{GraphHull} to recover multi-level community structure and to achieve competitive or superior performance in link prediction and community detection, while naturally providing interpretable predictions.
\end{abstract}

\section{Introduction}
In recent years, machine learning on graphs has attracted considerable attention.
This is not surprising since data from different domains can be modeled as graphs.
For instance, molecules~\citep{gilmer2017neural}, proteins~\citep{gligorijevic2021structure}, and computer programs~\citep{cheng2021deepwukong} are commonly represented as graphs.
Graph learning approaches mainly map elements of the graph into a low-dimensional vector space, while preserving the graph structure. 
A significant amount of research effort has been devoted to node embedding approaches, \ie algorithms that map the nodes of a graph into a low-dimensional space~\citep{cui2018survey}.
These methods typically use shallow neural network models to learn embeddings in an unsupervised manner. 
Those embeddings can then serve as input for various downstream tasks.
Graph neural networks (GNNs) are another family of graph learning algorithms which are mainly applied to supervised learning problems. 
These models follow a message passing procedure, where each node updates its feature vector by aggregating the feature vectors of its neighbors~\citep{wu2020comprehensive}.

Unsupervised learning and clustering methods play a central role in revealing latent structure and intrinsic organization in complex datasets. Among such approaches, \emph{Archetypal Analysis} (AA) \citep{cutler1994archetypal} provides a geometrically interpretable framework for representing data constrained to a $K$-dimensional polytope. In AA, each observation is expressed as a convex combination of extremal points, the \emph{archetypes} which correspond to the vertices of the polytope and capture pure, representative profiles of the data. Beyond its original formulation, AA has been studied extensively from both machine learning and geometric perspectives \citep{MORUP201254,damle2015geometricapproacharchetypalanalysis}, highlighting connections to matrix factorization and convex geometry. Its geometric interpretation has also provided insight into trade-offs and Pareto-optimal structure in biological systems \citep{shov}. Over the years, AA has been successfully applied to diverse domains such as computer vision \citep{6909588} and population genetics \citep{Gimbernat-Mayol2021.11.28.470296}, while significant effort has been devoted to improving scalability and robustness \citep{EUGSTER20111215,AA_prac}. A comprehensive overview of methodological developments and applications is provided in the recent survey of \citet{alcacer2025surveyarchetypalanalysis}. More recently, AA has been generalized to relational settings \citep{nakis2025signeda,nakis2023characterizing}, enabling applications to network data.

One prominent application of graph learning algorithms is community detection (a.k.a., clustering), \ie the problem of discovering clusters of nodes, with many edges joining nodes of the same cluster and comparatively few edges joining nodes of different clusters~\citep{fortunato2010community}.
In recent years, different graph learning algorithms have been proposed that can assign nodes to communities while maintaining high clustering quality~\citep{li2018community,shchur2019overlapping,chen2019supervised,sun2020network}.
Despite the success of these models in detecting communities in real-world networks, the lack of transparency still limits their application scope. 
Similar concerns have been raised in various domains, including healthcare and criminal justice~\citep{rudin2019stop}, and have led to the development of the field of explainable artificial intelligence (XAI) \citep{ribeiro2016should,ying2019gnnexplainer,gautam2022protovae}.
XAI builds methods that justify a model's predictions, thus increasing transparency, trustworthiness and fairness. However, existing approaches generally focus on optimizing clustering quality rather than providing explanations for the learned communities. 
Moreover, current explainable graph learning methods mostly target node classification tasks, leaving the problem of interpretable community detection relatively underexplored. To the best of our knowledge, no existing approach provides a principled generative framework that jointly achieves accurate detection of communities and inherent interpretability.

In this paper, we introduce \textsc{GraphHull}, a novel generative framework for 
explainable graph representation learning. The method relies on archetypes~\citep{cutler1994archetypal}, extreme profiles that summarize the diversity of structures found in a graph. Our contributions are: \textbf{i}) \emph{Hierarchical geometric design} by proposing a two-level convex-hull 
architecture in which global archetypes define extreme community profiles and 
anchor-dominant local hulls introduce interpretable prototypes for community-specific variation; 
\textbf{ii}) \emph{Identifiability by construction} in proving that local hulls are non-overlapping, which guarantees unique node reconstructions and transparent community assignments;
\textbf{iii}) \emph{Stable and scalable inference} by deriving a Lipschitz continuity 
result for the MAP objective, that enables provably safe optimization, and reducing likelihood evaluation from $\mathcal{O}(N^2)$ to $\mathcal{O}(|E|)$ via unbiased subsampling; 
\textbf{iv}) \emph{Principled diversity} through determinantal point process priors on both global and local archetypes, that encourage non-degenerate and expressive latent geometry; and 
\textbf{v}) \emph{Diverse empirical validation} via the competitive or superior performance of \textsc{GraphHull} on multiple real-world networks for link prediction and community detection, accompanied by interpretable multi-scale explanations.
Overall, our geometric design enforces non-overlapping convex hulls anchored at global archetypes, yielding community embeddings that are simultaneously unique, interpretable, and diverse. \textit{Code is available here:} \url{https://github.com/Nicknakis/GraphHull}

\section{Related Work}

\textbf{Graph learning for community detection.} 
Community detection has been approached using both shallow embedding methods and deeper models such as GNNs. 
The standard pipeline embeds nodes into a low-dimensional vector space and then applies clustering algorithms like $k$-means. 
Unsupervised node representation learning methods include random walk–based approaches, matrix factorization, and autoencoders. 
Although shallow embeddings were once considered less effective, methods such as DeepWalk~\citep{perozzi2014deepwalk} and node2vec~\citep{grover2016node2vec} have recently been shown to achieve the optimal detectability limit under the stochastic block model~\citep{kojaku2024network}. 
General-purpose embeddings often fail to capture community structure, motivating approaches that jointly preserve graph structure and optimize clustering objectives, \eg modularity~\citep{sun2020network}. 
Other works introduce community embeddings alongside node embeddings~\citep{cavallari2017learning}, or use nonnegative matrix factorization to directly encode community structure~\citep{li2018community}. GNNs have been applied to community detection, \eg with belief-propagation–inspired message passing~\citep{chen2019supervised,Wang_2025}, 
Bernoulli--Poisson models for overlapping communities~\citep{shchur2019overlapping}, 
and node classification formulations~\citep{wang2021unsupervised}. 
Autoencoder approaches embed nodes and then cluster them~\citep{yang2016modularity,tian2014learning,xie2019high}, 
often with GNN encoders~\citep{wang2017mgae,he2021community}, 
or unify embedding and clustering in a single framework~\citep{wang2019attributed,chen2019variational,choong2018learning}.

\textbf{Explainable graph learning algorithms.}
Graph explainability methods fall into two categories: post-hoc approaches and self-explainable models. 
Post-hoc methods explain trained GNNs by identifying influential subgraphs~\citep{ying2019gnnexplainer,yuan2020xgnn,vu2020pgm,luo2020parameterized,yuan2021explainability,lin2021generative,henderson2021improving}, 
but are sensitive to perturbations and thus less reliable~\citep{li2024explainable}. 
Self-explainable models are interpretable by design, often via motifs~\citep{lin2020graph,serra2024l2xgnn,miao2022interpretable} or prototypes~\citep{zhang2022protgnn,ragno2022prototype,dai2021towards}, with recent work formalizing their expressivity~\citep{chen2024interpretable} and synthetic data for benchmarking~\citep{agarwal2023evaluating}. 
Our approach is related to prototype-based methods, but instead learns \emph{archetypes}. 
In signed networks, archetypal embeddings have also been used for interpretable community structure~\citep{nakis2025signeda,nakis2025signedb}.

\section{Proposed Method}\label{sec:method}

\begin{figure*}[!h]
  \centering
    \includegraphics[width=1\linewidth]{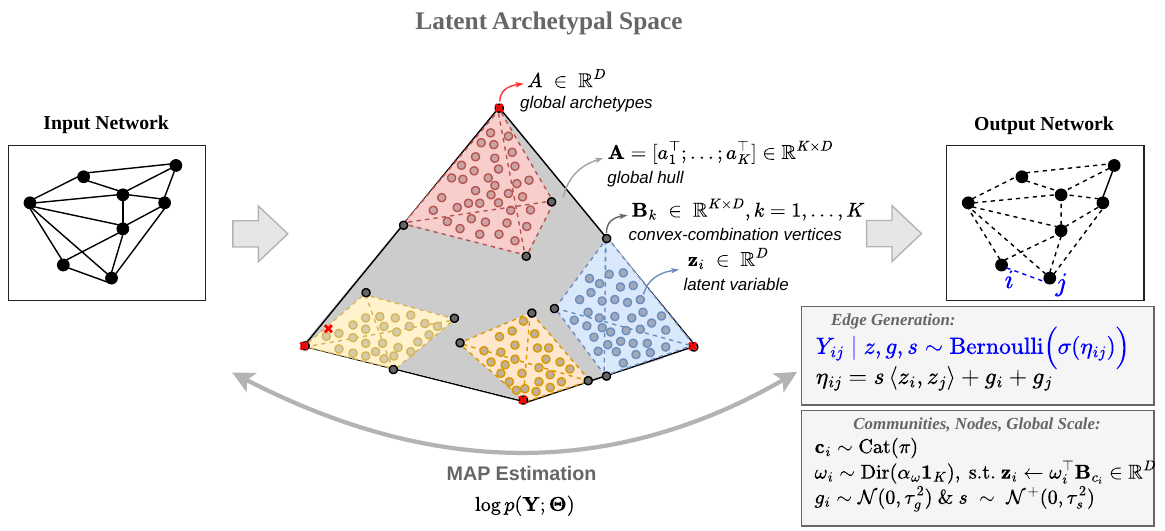}
  \caption{Visualization of the proposed archetypal graph generative model, so-called \textsc{GraphHull}.}
  \label{fig:model}
\end{figure*}

\textbf{Preliminaries.}
 Let \(\mathcal{G}=(V,E)\) be a simple undirected graph with \(N=|V|\) and adjacency matrix \(\bm{Y}\in\{0,1\}^{N\times N}\), where \(\bm{Y}_{ij}=\bm{Y}_{ji}\) and \(\bm{Y}_{ii}=0\). Capital bold letters such as \(\mathbf{X}\) denote matrices, small bold letters such as \(\mathbf{x}\) denote vectors, while non-bold letters such as \(x\) denote scalars. We next extend archetypal analysis to hierarchical nested structures by introducing a convex-hull formulation that captures latent structural organization in an inherently interpretable and generative manner, proposing \textsc{GraphHull}, illustrated in Figure \ref{fig:model}. A high-level generative process of the model is provided in Algorithm \ref{alg:gen} (the detailed generative process is provided in the supplementary material).

\textbf{Global archetypes.}
Given a graph $\mathcal{G}$, we aim to embed nodes in a $D$-dimensional latent space that is constrained to a $(D\!-\!1)$-dimensional polytope. Each node’s embedding is represented as a convex combination of the polytope’s vertices, where each vertex corresponds to an \emph{archetype} capturing a pure, extreme profile of the network \citep{nakis2023characterizing,nakis2025signeda}. Formally, we learn a matrix of archetypes
\begin{equation}
\bm{A} = [\bm{a}_1^\top;\dots;\bm{a}_K^\top] \in \mathbb{R}^{K\times D}\quad \text{with } K \leq D,
\end{equation}
where each row defines a corner of the polytope. This geometric characterization couples communities with extreme archetypal profiles, yielding interpretable latent factors that explain network structure. Such an archetypal analysis of the network can successfully capture global characteristics, but it disregards potential structure within each archetype, motivating the need for a multi-scale extremal organization of the network. 

A key design choice in our model is how to construct the global archetype matrix $\mathbf{A}$, which defines the vertices of the global convex hull in latent space. 
To ensure stability, we parameterize
$\mathbf{A}$ in an SVD-like form,
\begin{equation}
\label{eq:box_svd}
\bm{A} \;=\; \bm{U} \,\mathrm{diag}(\bm{\sigma})\, \bm{V}^\top,
\end{equation}

where $\bm{U}\in\mathbb{R}^{K\times K}$ and $\bm{V}\in\mathbb{R}^{D\times K}$ have
orthonormal columns, and $\bm{\sigma}\in\mathbb{R}^K$ contains the singular values constrained to lie
within fixed bounds $[\sigma_{\min},\sigma_{\max}]$.
This construction has three advantages:
(i) it produces archetypes that span approximately orthogonal directions,
reducing the risk of degeneracy;
(ii) the bounded singular values control the overall scale of inner products,
preventing either collapse or explosion of the latent embeddings; and
(iii) the separation of orientation (via $\bm{U},\bm{V}$) and scale (via $\sigma$)
reduces redundant variability and stabilizes inference. Enforcing $\sigma_{\min}>0$ guarantees that the rows of
$\mathbf{A}$ are linearly independent, so that $\mathbf{A}$ always spans a
well–conditioned latent basis.

\begin{algorithm}[!t]
\scriptsize
\caption{Generative process}
\label{alg:gen}
\begin{algorithmic}[1]
\State \textbf{Global archetypes:} Sample $\mathbf{A}\in\mathbb{R}^{K\times D}$
       using the boxed SVD parameterization.
\State \textbf{Local hulls:} For each community $k=1,\dots,K$:
       construct prototypes $\mathbf{B}_k=\widetilde{\mathbf{W}}_k\mathbf{A}$
       with anchor–dominant rows (truncated Dirichlet, $\varepsilon$).
       If $\varepsilon<\tfrac12$, then $\conv(\mathbf{B}_k)\cap \conv(\mathbf{B}_\ell)=\varnothing$.
\State \textbf{Diversity priors:} Place DPP priors on rows of $\mathbf{A}$ and
       each $\mathbf{B}_k$ to encourage spread.
\State \textbf{Communities and nodes:}
       Draw community proportions $\bm{\pi}$, assignments $c_i\sim\text{Cat}(\bm{\pi})$,
       barycentric weights $\bm{\omega}_i\sim\text{Dir}(\alpha_\omega)$,
       and set $\bm{z}_i=\bm{\omega}_i^\top \mathbf{B}_{c_i}$.
       Draw degree bias $g_i\sim\mathcal{N}(0,\tau_g^2)$.
\State \textbf{Global scale:}
       Draw $s \sim \mathrm{HalfNormal}(\tau_s)$.
\State \textbf{Edges:} For each $i<j$, sample
       $Y_{ij}\sim \text{Bern}(\sigma(s\, \bm{z}_i^\top \bm{z}_j+g_i+g_j))$.
\end{algorithmic}
\end{algorithm}

\textbf{Anchor–dominant local hulls.}
We posit $K$ \emph{global archetypes} collected in 
$\mathbf{A}\in\mathbb{R}^{K\times D}$, each serving as an 
extremal profile that defines a latent community in the network. To capture community–specific structure at a finer resolution, for each vertex (archetype) of the global hull $k\in\{1,\dots,K\}$ we introduce a 
\emph{local prototypical hull} $\mathbf{B}_k\in\mathbb{R}^{K\times D}$.
Each node will be represented as a convex combination of the \(K\) rows of some \(\mathbf{B}_k\), yielding interpretable community‑aware prototypes.
The idea parallels prototype‑based modeling in interpretable image generation \citep{gautam2022protovae,kjaersgaard2024pantypes}, but here it is adapted to relational data where both global and local structure must be inferred jointly.

At the global level, $\mathbf{A}$ contains extremal archetypes that
define pure community profiles. At the local level, each hull
$\mathbf{B}_k$ refines a community by introducing vertices that act
as prototypes/local archetypes, i.e., representative subprofiles within the community. For enhanced identifiability, 
each local hull \(\mathbf{B}_k\) is \emph{anchored} to a distinct global archetype \(\bm{a}^{\top}_k\).
We write
\begin{equation}
\label{eq:constr}
    \mathbf{B}_k
\;=\; \Tilde{\bm{W}}_k\bm{A}=
\begin{bmatrix}
\mathbf{W}_k\\[0.25ex]
\bm{e}_k^\top
\end{bmatrix}\mathbf{A}
\;\in\;\mathbb{R}^{K\times D},
\qquad
k=1,\dots,K,
\end{equation}
where the total barycentric coordinates (including the anchor) are given by $\Tilde{\bm{W}}_k$ so that \(\bm{e}_k^\top\in\mathbb{R}^{1\times K}\) is the \(k\)-th standard basis row (so the last row of \(\mathbf{B}_k\) equals the anchor \(\bm{a}_k^\top\)), and
\(\mathbf{W}_k\in\mathbb{R}^{(K-1)\times K}\) is a \emph{community‑specific} row–simplex matrix whose rows are convex weights over the global archetypes:
\[
(\mathbf{W}_k)_{\ell,:}\in\Delta_{K-1}
\;\;:=\;\{\,\bm{w}\in\mathbb{R}^K:\; \bm{w}\ge 0,\;\mathbf{1}^\top \bm{w}=1\,\},
\]
\[\qquad \ell=1,\dots,K-1.\]

Under a generative model, for selecting the anchor, one may regard this as sampling a global archetype without replacement and fixing it as a vertex of the local hull.
The remaining \(K-1\) vertices of \(\mathbf{B}_k\) are constrained to be convex combinations of the global archetypes, ensuring that local extremes are explained hierarchically by the global structure, as prototypes.
Thus, \(\mathbf{B}_k\) has \(K\) rows (vertices in \(\mathbb{R}^D\)): one fixed anchor at the global archetype \(\bm{a}_k^{\top}\) and \(K-1\) prototypes with rows obtained as convex combinations of \(\{\bm{a}^{\top}_j\}_{j=1}^K\).
Importantly, \(\mathbf{B}_1,\dots,\mathbf{B}_K\) are \emph{distinct} (each has its own coefficient matrix \(\mathbf{W}_k\)) and are anchored to different global archetypes.


\textbf{Non–overlapping local hulls by construction.}
A central property of our design is \emph{local hull identifiability}.
Therefore, for each pair of communities $k\neq \ell$, their convex hulls should satisfy
$\mathrm{conv}(\mathbf{B}_k)\;\cap\;\mathrm{conv}(\mathbf{B}_\ell)\;=\;\varnothing.$
This disjointness ensures that every node embedding has a unique reconstruction:
if two local hulls were to overlap, then any point in the intersection could be expressed as a convex combination of prototypes from either hull, breaking identifiability and interpretability of the representation. While one could attempt to discourage overlap by adding prior encouraging hull separation, these provide only a soft penalty and cannot guarantee non–overlap without careful tuning.
Instead, we ensure disjointness \emph{by construction} by defining the local hulls
\(\mathbf{B}_k\) through barycentric coordinates \(\mathbf{W}_k\) sampled from a
\emph{truncated Dirichlet prior}. Specifically, each row \(\bm{w}^{\top}\) is drawn as
$\bm{w} \;\sim\; \mathrm{Dir}(\alpha \mathbf{1}_K) \quad
\text{conditioned on } w_k \;\ge\; 1-\varepsilon,$
where \(k\) denotes the anchor coordinate of hull \(k\).
This truncation forces every local prototype in the hull anchored at the \(k\)-th global archetype (\(\bm{a}^{\top}_k\)) to retain at least mass \(1-\varepsilon\) on its anchor.
For \(\varepsilon < \tfrac{1}{2}\) this ensures that anchor-dominant regions of different
hulls are disjoint, so the induced convex hulls do not overlap cf.\ Lemma~\ref{lem:nonoverlap}, (proof in supplementary).

\begin{lemma}[Non-overlap of anchor-dominant local hulls]
\label{lem:nonoverlap}
Let $\bm{A}\in\mathbb{R}^{K\times D}$ be affinely independent global archetypes.
For each community $k\in\{1,\dots,K\}$, let its local barycentric rows
$\bm{w}\in\Delta_{K-1}$ be drawn from a truncated Dirichlet prior
\[
\bm{w} \sim \mathrm{Dir}(\alpha \mathbf{1}_K) \;\;\text{conditioned on } w_k \ge 1-\varepsilon,
\quad 0<\varepsilon<\tfrac{1}{2}.
\]
Let $\Tilde{\bm{W}}_k$ stack $K$ such rows, with the final row set to $\bm{e}^{\top}_k$, and define local
vertices $\bm{B}_k=\Tilde{\bm{W}}_k \bm{A}$. Then for any $k\neq \ell$,
\[
\mathrm{conv}(\bm{B}_k)\ \cap\ \mathrm{conv}(\bm{B}_\ell)\;=\;\varnothing.
\]
\end{lemma}
In practice we realize this truncated Dirichlet prior by parameterizing each
row of $\bm{W}_k\in\mathbb{R}^{(K-1)\times K}$ directly. Let $\varepsilon\in(0,\tfrac12)$
be the anchor mass. For rows $r=1,\dots,K-1$ we set
\begin{align}
\label{eq:anchor-dominant}
(\mathbf{W}_k&)_{r,:} = (1-s_{k,r})\,\bm{e}^{\top}_k + s_{k,r}\,\Tilde{ \bm{q}}_{k,r}, \\ \nonumber
&(\mathbf{W}_k)_{r,:}\in\Delta_{K-1},\;\;  s_{k,r} = \varepsilon\, t_{k,r}, \;\; t_{k,r}\in(0,1),
\end{align}
where $\bm{e}_k$ is the $k$-th basis vector and
$\Tilde{\bm{q}}_{k,r}\in\Delta_{K-1}$ is a probability vector supported only on
$\{1,\dots,K\}\setminus\{k\}$ (i.e.\ zero mass on the anchor coordinate) where we can a place Dirichlet
prior, $\Tilde{\bm{q}}_k \;\sim\; \mathrm{Dir}\!\left(\alpha_{q}\,\mathbf{1}_K\right)$.
The final row is set to $\bm{e}^{\top}_k$, yielding $\Tilde{\bm{W}}_k\in\mathbb{R}^{K\times K}$,
and the local vertices are obtained as $\mathbf{B}_k=\Tilde{\bm{W}}_k \bm{A}\in\mathbb{R}^{K\times D}$.
This anchor–dominant parameterization is equivalent to sampling from the truncated Dirichlet in Lemma~\ref{lem:nonoverlap}, but is simpler to implement in practice and ensures the disjointness guarantee. Finally, the strength of anchor dominance in Eq. \eqref{eq:anchor-dominant} is
controlled by the shrink variable $s_{k,r}=\varepsilon\,t_{k,r}$, where
$t_{k,r}\in(0,1)$ is given a Beta prior $t_{k,r} \;\sim\; \mathrm{Beta}(a,b).$ 


\textbf{Node representation.}
Having the global and local archetypal structure in place we can define the final node representation. For each node, we define a
global community/archetype assignment to exactly one community (local hull). We encode this with the one–hot matrix,
\(\mathbf{M}\in\{0,1\}^{N\times K}\). Conceptually, each row \(\bm{m}^{\top}_i\) corresponds to drawing the assignment $c_{i}$ from a categorical distribution over the \(K\) global archetypes, so that \(\bm{m}^{\top}_i\) is a one–hot vector indicating the unique hull to which node \(i\) belongs. 
If node \(i\) belongs to community \(c_i\), then \(\bm{m}^{\top}_i=\bm{e}^{\top}_{c_i}\) is the $i$-th row of \(\mathbf{M}\).
Within its community, the position of the node is given by barycentric coordinates
\(\bm{\omega}_i \in \Delta_{K-1}\) over the local archetypes \(\mathbf{B}_{c_i}\in\mathbb{R}^{K\times D}\).
Thus the embedding of node \(i\) is simply
\begin{equation}
\bm{z}_i \;=\; \bm{\omega}_i^\top\, \mathbf{B}_{c_i} \;\in\;\mathbb{R}^D.
\label{eq:zi}
\end{equation}

Stacking all nodes yields the final representation matrix
\(\mathbf{Z}=[\bm{z}_1^\top;\dots;\bm{z}_N^\top]\in\mathbb{R}^{N\times D}\) while \(\Omega=[\bm{\omega}_1^\top;\dots;\bm{\omega}_N^\top]\in\mathbb{R}^{N\times K}\) collects the barycentric weights. For the node–specific barycentric weights $\bm{\omega}_i$ over the $K$ local
archetypes of its assigned hull $\mathbf{B}_k$, we place a symmetric Dirichlet
prior, $\bm{\omega}_i \;\sim\; \mathrm{Dir}\!\left(\alpha_{\omega}\,\mathbf{1}_K\right),$
with $\alpha_{\omega}=1$, corresponding to the uniform distribution over the
probability simplex. 
During training we employ the Gumbel–Softmax (GS) relaxation \citep{jang2017categorical}, over global assignment matrix \(\mathbf{M}\in\{0,1\}^{N\times K}\) which provides a differentiable approximation to such categorical samples.


\textbf{Edge likelihood.}
We focus on binary undirected graphs and model edges with a Bernoulli likelihood. The formulation can be extended to weighted graphs using a Poisson likelihood and to signed graphs using a Skellam likelihood. Let $g_i\in\mathbb{R}$ be a node-specific degree bias, $\bm{z}_i\in\mathbb{R}^D$ the latent representation of node $i$, and $s>0$ a global scale parameter controlling geometric sharpness. For an unordered node pair $(i,j)$ with $i<j$, the log-odds expression is
\begin{align}
\eta_{ij}
&= s\,\langle \bm{z}_i, \bm{z}_j\rangle \;+\; g_i+g_j, \label{eq:linpred}\\
&Y_{ij}\mid \bm{Z},\bm{g},s \sim \mathrm{Bernoulli}\big(\sigma(\eta_{ij})\big). \nonumber
\end{align}
The complete log-likelihood over the set of unordered pairs
$\mathcal{D}=\{(i,j):\,1\le i<j\le N\}$ is
\begin{equation}
\label{eq:loglik}
\log p(\bm{Y}\mid \bm{Z},\bm{g},s)=\sum_{i<j} \Big[\, Y_{ij}\,\eta_{ij} - \log(1+\exp\eta_{ij}) \,\Big].
\end{equation}

We place independent Gaussian priors on the node biases,
$g_i \;\sim\; \mathcal{N}(0,\tau_{i,g}^2)$,
while for the global scale parameter $s>0$, we use a half-normal prior $s \;\sim\; \mathcal{N}^+(0,\tau_s^2)$.

\textbf{Diversity of vertices within each local hull.}
An important characteristic of our model is the \emph{diversity and expressiveness}
of both the local hulls $\mathbf{B}_k$ and global hull $\bm{A}$.
To capture rich community structure, each hull should span a large and
well–conditioned volume, rather than collapsing onto a small subspace or producing
co–linear vertices.
To encourage this, we place a \emph{Determinantal Point Process (DPP)} \citep{kulesza2012determinantal}
prior in its $L$–ensemble formulation on the set of local/global archetypes. The $L$–ensemble DPP requires a positive–semidefinite kernel matrix.

For a convex hull, let $\bm{\Phi}\in\mathbb{R}^{K\times D}$ denote its $K$ vertices
and define the row–normalized matrix $\bm{\Psi}$, with rows
$\bm{\psi}_{k}=\bm{\phi}_{k}/\|\bm{\phi}_{k}\|_2$, we then form the Gram matrix $\bm{L}=\bm{\Psi} \bm{\Psi}^\top$. The log–prior contribution is
\begin{equation}
\label{eq:dpp}
\log p_{\mathrm{DPP}}(\bm{\Phi})
\;=\;
\log\det(\bm{L})\;-\;\log\det(\bm{I}+\bm{L}).
\end{equation}

This determinantal point process prior acts as a repulsive force among both the local and global archetypes, discouraging degeneracy and collapse. Applied to the local hulls $\mathbf{B}_k$, it encourages the vertices to be diverse so that each hull spans a rich volume within its community. Applied to the global archetypes $\mathbf{A}$, it spreads the global basis vectors apart, ensuring that the communities themselves are well separated.

\textbf{Joint distribution and MAP objective.} Let $\Theta$ denote the \textsc{GraphHull} parameters,
with deterministic constraints $\mathbf{B}_k=\Tilde{\bm{W}}_k\mathbf{A}$ and $\bm{z}_i$ given by Eq. \eqref{eq:zi}.
Up to constants, the joint distribution factorizes as:
\vspace{-2pt}
\begin{equation}
\begin{aligned}
&\log p(\bm{Y},\bm{\Theta})
= \underbrace{\sum_{(i,j)\in\mathcal{D}} 
    \Big[Y_{ij}\,\eta_{ij} - \mathrm{softplus}(\eta_{ij})\Big]}_{\text{Bernoulli edge likelihood}} +\\
& \underbrace{\sum_{i=1}^{N}
    (\alpha_{\bm{\omega}}-1)\!\sum_{k=1}^{K}\log \omega_{ik}}_{\text{Dirichlet prior on }\omega_i}  + \underbrace{\sum_{k=1}^{K}\sum_{r=1}^{K-1} (\alpha_{q}-1)\!\!\sum_{j\ne k}\log (q_{k,r})_j}_{\text{non-anchor Dirichlet}}  \\
& + \underbrace{\sum_{k=1}^{K}\sum_{r=1}^{K-1}
    \Big[(a-1)\log t_{k,r} + (b-1)\log (1-t_{k,r}) \Big]}_{\text{anchor shrinkage (Beta)}} \\
&+ \underbrace{\sum_{k=1}^{K}
    \Big[\log\det(\bm{L}_k)-\log\det(\bm{I}+\bm{L}_k)\Big]}_{\text{Local hull DPP prior }} \\
& + \underbrace{
    \Big[\log\det(\bm{L})-\log\det(\bm{I}+\bm{L})\Big]}_{\text{Global Hull DPP prior}} - \frac{1}{2\tau_g^2}\sum_{i=1}^N g_i^2 - \frac{s^2}{2\tau_s^2}\, .
\end{aligned}
\label{eq:joint}
\end{equation}

 Naively, evaluating the Bernoulli–logistic edge likelihood scales as \(\mathcal{O}(N^2)\) due to all pairwise interactions. To achieve scalability, we exploit that the first term involves only observed edges and can be computed in \(\mathcal{O}(|E|)\), while the log-partition term is estimated by uniform subsampling of non-edges, yielding an unbiased estimator. This reduces the per-iteration cost to \(\mathcal{O}(|E|)\). The DPP prior adds a smaller overhead. For each local hull \(\bm{B}_k \in \mathbb{R}^{K\times D}\) (and the global archetypes \(A\)), we compute a Gram matrix and two log-determinants, which require \(\mathcal{O}(K^2D + K^3)\) operations. Since there are \(K\) local hulls plus one global term, the total DPP cost is \(\mathcal{O}(K^3 D + K^4)\). The remaining priors contribute lower-order costs. Thus, the overall per-iteration complexity of the method is $\mathcal{O}(|E|) + \mathcal{O}(K^3 D + K^4)$, dominated by the linear-in-edges term for large graphs when \(K \ll N\).

 \textbf{Optimization and curvature.}
Boxing the global archetype matrix not only improves identifiability but also controls curvature: the next theorem shows that the gradient with respect to the barycentric weights is globally Lipschitz, so fixed-step projected gradient is safe, as shown in Theorem \ref{thm:boxing-lipschitz} (proof in the supplementary). This helps explain the stable behavior we observe in practice.

\begin{theorem}[Boxing $\Rightarrow$ Lipschitz gradient in $\boldsymbol{\omega}$]
\label{thm:boxing-lipschitz}
Consider the negative Bernoulli--logistic loss
\begin{align}
\mathcal{L}(\bm{Z},\bm{g},s)=&\sum_{i<j}\big[\log( 1 +\exp \eta_{ij})-Y_{ij}\eta_{ij}\big]. 
\end{align}
Assume (i) the archetype matrix is boxed, $\|\mathbf{A}\|_2\le \kappa_\star$; and
(ii) each embedding $\mathbf{z}_i$ is a convex combination of rows of $\mathbf{A}$
(via local hull vertices), hence $\|\mathbf{z}_i\|\le \kappa_\star$.
Let $\boldsymbol{\omega}=(\boldsymbol{\omega}_1,\dots,\boldsymbol{\omega}_N)$
be the concatenated barycentric weights. Then the gradient
$\nabla_{\boldsymbol{\omega}}\mathcal{L}$ is Lipschitz with constant
\[
L_{\omega} \;\le\; \frac{1}{4}\left( s\,\kappa_\star^{2}\,\deg_{\max}
~+~ \tfrac{s^2}{2}\,\kappa_\star^{4}\,\deg_{\max}\right),
\]
where $\deg_{\max}\le N-1$ is the maximum degree.
Consequently, projected gradient on the product of simplices with any
stepsize $\eta \le 1/L_{\omega}$ is globally safe (monotone descent).
\end{theorem}

\noindent\textbf{Remark (MAP).}
Adding the smooth priors in Eq.~\eqref{eq:joint} preserves Lipschitz continuity:
the total MAP gradient is Lipschitz with constant at most
$L_{\omega,\text{MAP}} \le L_{\omega} + L_{\omega,\text{prior}}$, where
$L_{\omega,\text{prior}}$ is finite for Gaussian and log-determinantal terms. The Lipschitz bound depends on $s>0$, the global scale parameter.
In practice, $s$ is regularized by a half-normal prior in the MAP objective keeping the Lipschitz constant finite.

\textbf{Interpretability and Self-Explainability.}
Following the literature on self-explainable models (SEM) \citep{gautam2022protovae}, we characterize \textsc{GraphHull} through complementary notions of interpretability and explainability. 
The model is \emph{interpretable} because its latent components, global archetypes and anchor-dominant local prototypes organized as nested convex hulls, correspond to extremal structural roles in the graph. 
It is \emph{explainable} because each prediction admits an explicit generative decomposition: every node has a unique barycentric representation within a local hull, communities are anchored at global archetypes, and edge log-odds decompose into interpretable geometric interactions and degree effects. 
Thus, explanations arise directly from the model's latent geometry rather than post-hoc analysis. Moreover, \textsc{GraphHull} satisfies the core SEM predicates of \emph{transparency}, \emph{diversity}, and \emph{trustworthiness} \citep{gautam2022protovae}. 
Transparency follows from the direct use of archetypes and prototypes in the generative mechanism. 
Diversity is enforced via affine independence, anchor-dominant disjoint hulls, and determinantal point process priors. 
Trustworthiness stems from faithful generative decompositions and the boxed-SVD parameterization, which yields stable optimization with controlled curvature. Finally, as a generative SEM, \textsc{GraphHull} enjoys \emph{generative consistency}: the same interpretable latent factors that explain predictions also govern the probabilistic data-generation process. 
The hierarchical identifiability of prototypes through global archetypes further strengthens this alignment between geometry, explanation, and generation.

\section{Results}
We extensively evaluate \textsc{GraphHull} against baseline graph representation learning methods on networks of varying sizes and structures. 
All experiments were conducted on an 8\,GB Apple M2 machine. 
For \textsc{GraphHull}, we optimize the MAP objective of Eq.~\eqref{eq:joint} using Adam~\citep{kingma2014adam} with learning rate in range $[0.01,0.05]$. 
Unless otherwise stated, we set the anchor strength to $\varepsilon=0.45$, box SVD bounds to $[\sigma_{\min}=0.3,\sigma_{\max}=1.5]$, and choose $K=D$ for simplicity. Detailed description of hyperparameters for all models are provided in the supplementary.

\begin{table}[!t]
\begin{center}
\caption{Statistics of networks. $N$: number of nodes, $|E|$: number of edges, $K$ number of communities.}
\label{tab:network_statistics}
\resizebox{0.45\textwidth}{!}{%
 \begin{tabular}{rccccccccc}\toprule
 &\textsl{LastFM}&\textsl{Citeseer} & \textsl{Pol}& \textsl{Cora} & \textsl{Dblp} & \textsl{AstroPh} & \textsl{GrQc} &  \textsl{HepTh} \\\midrule
$N$&7,624&3,327 &18,500& 2,708 & 27,199 & 17,903 & 5,242  & 8,638 \\
$|E|$&55,612&9,104 &61,200 & 5,278 & 66,832 & 197,031 & 14,496 & 24,827  \\
$K$&14&6 &2& 7 & --- & --- & --- & --- \\\bottomrule
\end{tabular}%
}
\end{center}
\end{table}

\begin{table*}[!t]
\caption{AUC ROC scores for representation sizes of $8$, $16$, $32$, and $64$ averaged over five runs.}
\label{tab:AUC}

\centering
\resizebox{0.9\textwidth}{!}{%
\begin{tabular}{l*{20}{c}}
\toprule
 & \multicolumn{4}{c}{\textsl{AstroPh}}
 & \multicolumn{4}{c}{\textsl{GrQc}}
 & \multicolumn{4}{c}{\textsl{HepTh}}
 & \multicolumn{4}{c}{\textsl{Cora}}
 & \multicolumn{4}{c}{\textsl{DBLP}} \\
\cmidrule(lr){2-5}\cmidrule(lr){6-9}\cmidrule(lr){10-13}\cmidrule(lr){14-17}\cmidrule(lr){18-21}
\multicolumn{1}{l}{Dimension ($D$)} &
  8 & 16 & 32 & 64 &
  8 & 16 & 32 & 64 &
  8 & 16 & 32 & 64 &
  8 & 16 & 32 & 64 &
  8 & 16 & 32 & 64 \\
\midrule
\textsc{Node2Vec}    &.943  & \uline{.954} &\uline{.961}  &\uline{.962}  &\textbf{.928}  &\uline{.932}  &\uline{.937}  &\uline{.936}  & \textbf{.879} &.882  & .888 & .892 & \uline{.761} & .760 &\uline{.766}  & \uline{.777} & .920 & .923 & .931 & .941 \\
  \textsc{Role2Vec} & \textbf{.957}  &\textbf{.969}   &\textbf{.970}   & \textbf{.965}  & \uline{.927}  &\textbf{.936}    &.934  &.934  &\uline{.897}  & \textbf{.907} & \textbf{.902}  & \uline{.895}  & \textbf{.769} &\uline{.767} & .759  &.752  & \textbf{.940} &\textbf{.952}  & \uline{.943} & \uline{.944}  \\
\textsc{NetMF}       &.904  &.928  &.946  &.955  &.835  &.882  &.882  &.883  & .778 & .797 & .802 & .793 & .698 &.675  &.674  &.654  &.791  & .817 &.829  &.842  \\
\textsc{GraRep}  & .919 &.946  & .959 & \textbf{.965} & .892  &.906   & .909  &  .894 & .825  & .845  & .842 &.829  & .692 &.704  &.728  &.717  & .855 & .877 &.879  &.867  \\
\textsc{RandNE}      &.867  &.888  &.900  &.907  &.787  &.826  &.854  &.877  & .718 &.770  &.812  &.839  &.604  & .641 &.676  & .701 & .741 & .795 &.836  & .865 \\
\midrule
\textsc{NNSED}   &.863  &.883  &.891  &.897  &.799  &.836  &.861  & .876 &.741  &.788  & .820 &.841  &.620  &.649  &.689  &.709  &.755  &.801  &.834  &.853  \\
\textsc{MNMF}       & .877 & .916 & .938 &.954  & .881 &.905  &.918  &.918  &.810  &.844  & .864 &.875  &.694  &.718  &.700  &.680  &.859  &.898  &.919  & .931 \\
\textsc{SymmNMF}       & .784 & .818 & .840 & .862 &.859  &.873  &.901  & .905 & .740 &.770  &.795  &.826  &.734  & .717 &.713  & .705 & .822 &.848  &.868  &.888  \\
\bottomrule
\textsc{Dmon}       & .845 &.852 & .857&   .861 & .822 & .845 &.856  & .866 &.786 &.799 & .804 &.807  &.723  & .744 & .777 & .788 & .774 & .785 &.788  &.791  \\
\midrule
\textsc{GraphHull} & \uline{.945} & \uline{.954} &.960  &\textbf{.965}  &.923  &.931  &\textbf{.940}  &\textbf{.945}  & .874 & \uline{.885} & \uline{.895} & \textbf{.904} &.753  & \textbf{.790} & \textbf{.791} & \textbf{.803} &\uline{.930}  &\uline{.934}  &\textbf{.945}  & \textbf{.948} \\
\bottomrule
\end{tabular}%
}
\end{table*}

\textbf{Datasets and Baselines.} We evaluate on multiple undirected citation, collaboration, and social networks, each treated as unweighted (see Table~\ref{tab:network_statistics} for details). 
Citation networks include \textsl{Cora} (machine learning publications, $7$ classes) and \textsl{Citeseer} (computer science publications, $6$ classes) \citep{cora}. 
Collaboration networks include \textsl{DBLP} (co-authorship by research field) \citep{dblp}, \textsl{AstroPh} (astrophysics), \textsl{GrQc} (general relativity and quantum cosmology), and \textsl{HepTh} (high-energy physics) \citep{leskovec2007graph}. 
We also use the social network \textsl{LastFM} (Asian users with $14$ country labels) \citep{lastfm}, and the socio-political retweet network \textsl{Pol} (labels denote political alignment) \citep{nr}. We compare against shallow embeddings, matrix factorization methods, mixed-membership models, and a GNN-based approach. 
Shallow embeddings include \textsc{Node2Vec}~\citep{node2vec} and \textsc{Role2Vec}~\citep{role2vec}. 
Factorization-based methods are \textsc{NetMF}~\citep{netmf}, \textsc{GraRep}~\citep{grarep}, and \textsc{RandNE}~\citep{randne}. 
Mixed-membership models include \textsc{NNSED}~\citep{NNSED}, \textsc{MNMF}~\citep{MNMF}, and \textsc{SymmNMF}~\citep{SymmNMF}. 
We also evaluate \textsc{Dmon}~\citep{tsitsulin2023graphclusteringgraphneural}, which couples a GNN encoder with modularity maximization (for extended details see supplementary). We emphasize shallow embeddings, since they were recently shown to reach the detectability limit under the stochastic block model~\citep{kojaku2024network}.


\begin{figure*}[!t]
\centering
\begin{subfigure}{0.49\textwidth}  
    \centering
    \includegraphics[width=0.8\linewidth]{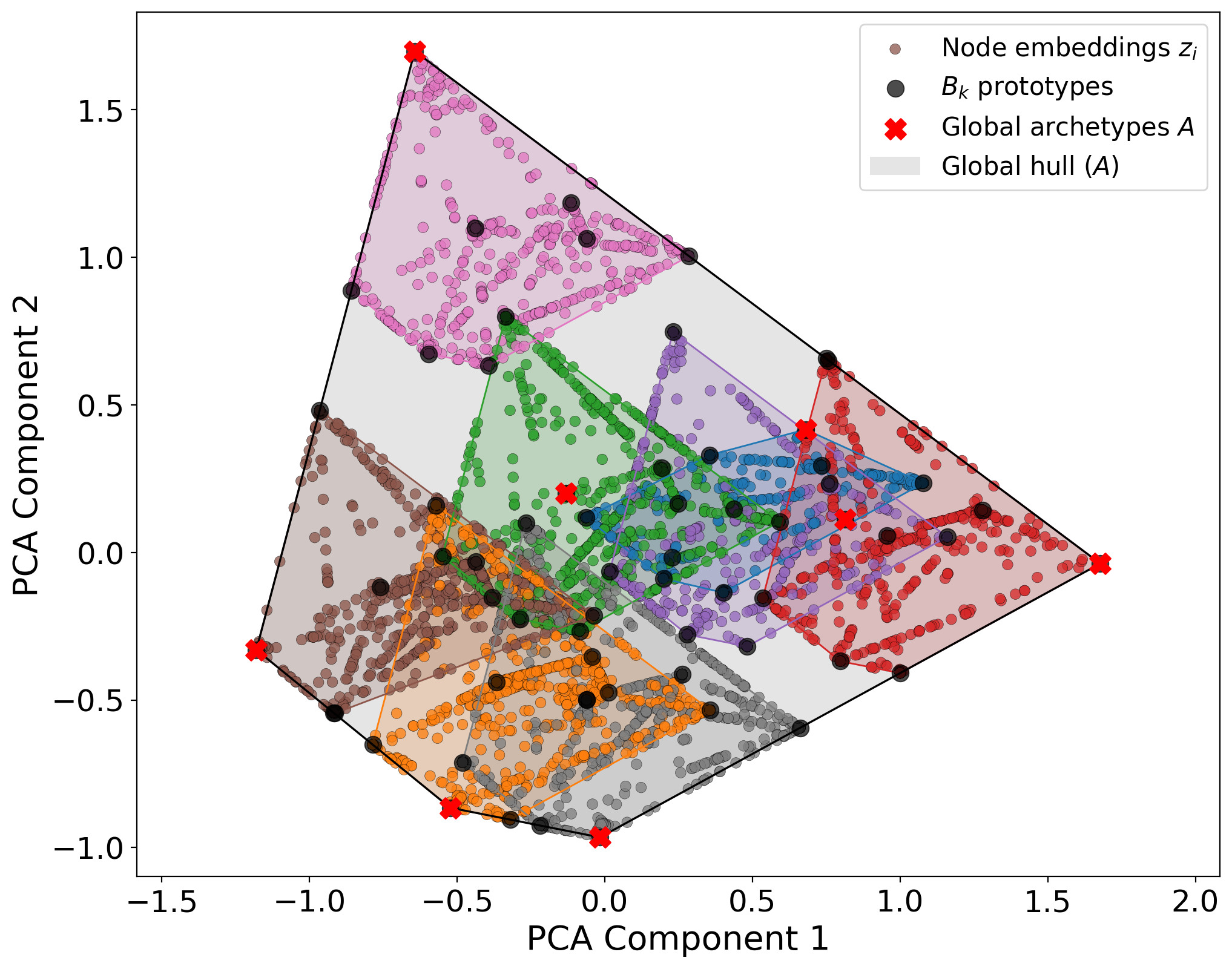}
    \caption{\textsc{GraphHull} PCA projection with $K=8$ and $D=8$.}
    \label{fig:1_level_pol_1}
\end{subfigure}
\hfill
\begin{subfigure}{0.49\textwidth}  
    \centering
    \includegraphics[width=0.61\linewidth]{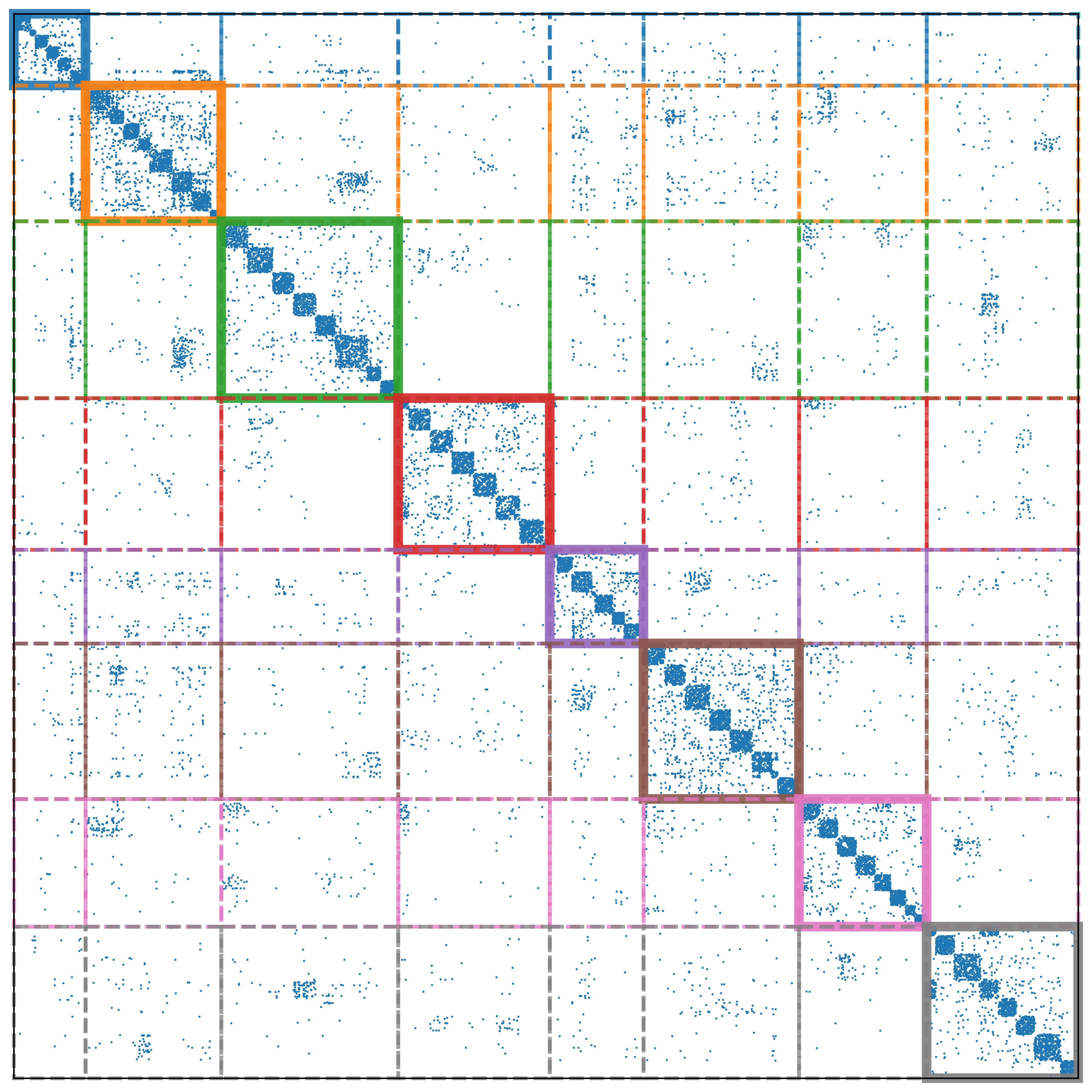}
    \caption{ \textsc{GraphHull} Reordered Adjacency matrix.}
    \label{fig:1_level_pol_2}
\end{subfigure}

\centering
\begin{subfigure}{0.11\textwidth}  
    \centering
    \includegraphics[width=1\linewidth]{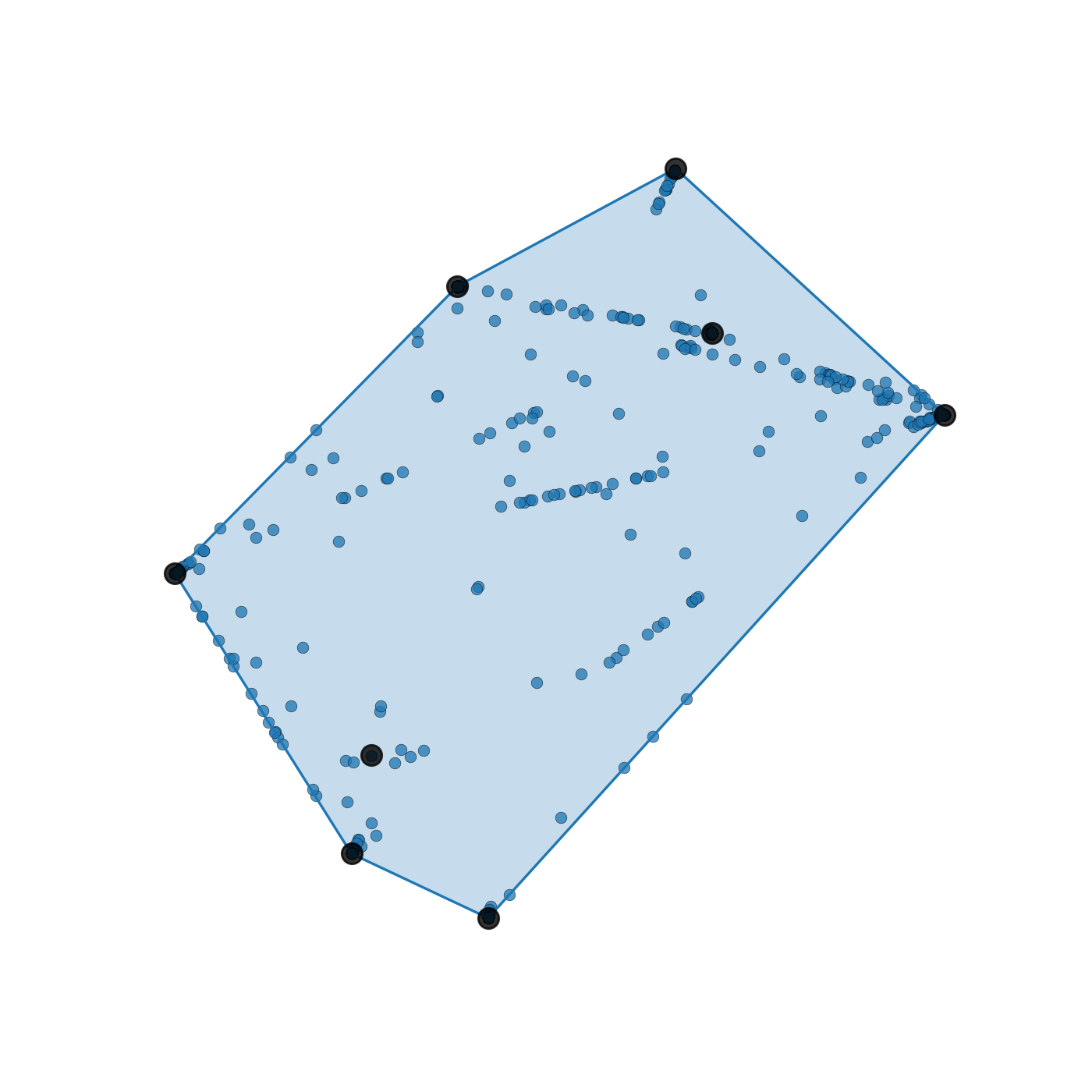}
    \caption{PCA: $\bm{B}_1$}
    \label{fig:local_hull_0}
\end{subfigure}
\hfill
\begin{subfigure}{0.11\textwidth}  
    \centering
    \includegraphics[width=1\linewidth]{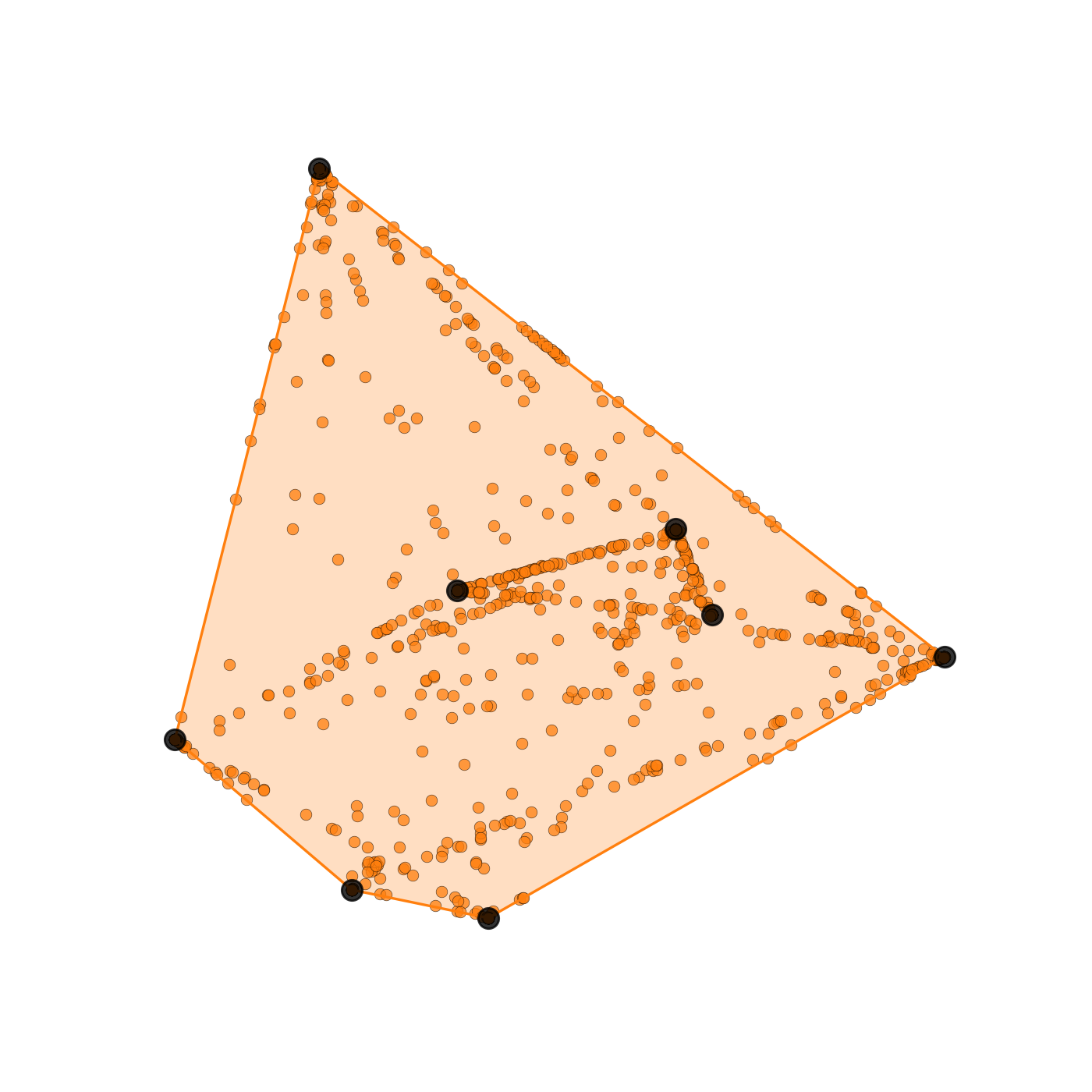}
    \caption{PCA: $\bm{B}_2$}
    \label{fig:local_hull_1}
\end{subfigure}
\hfill
\begin{subfigure}{0.11\textwidth}  
    \centering
    \includegraphics[width=1\linewidth]{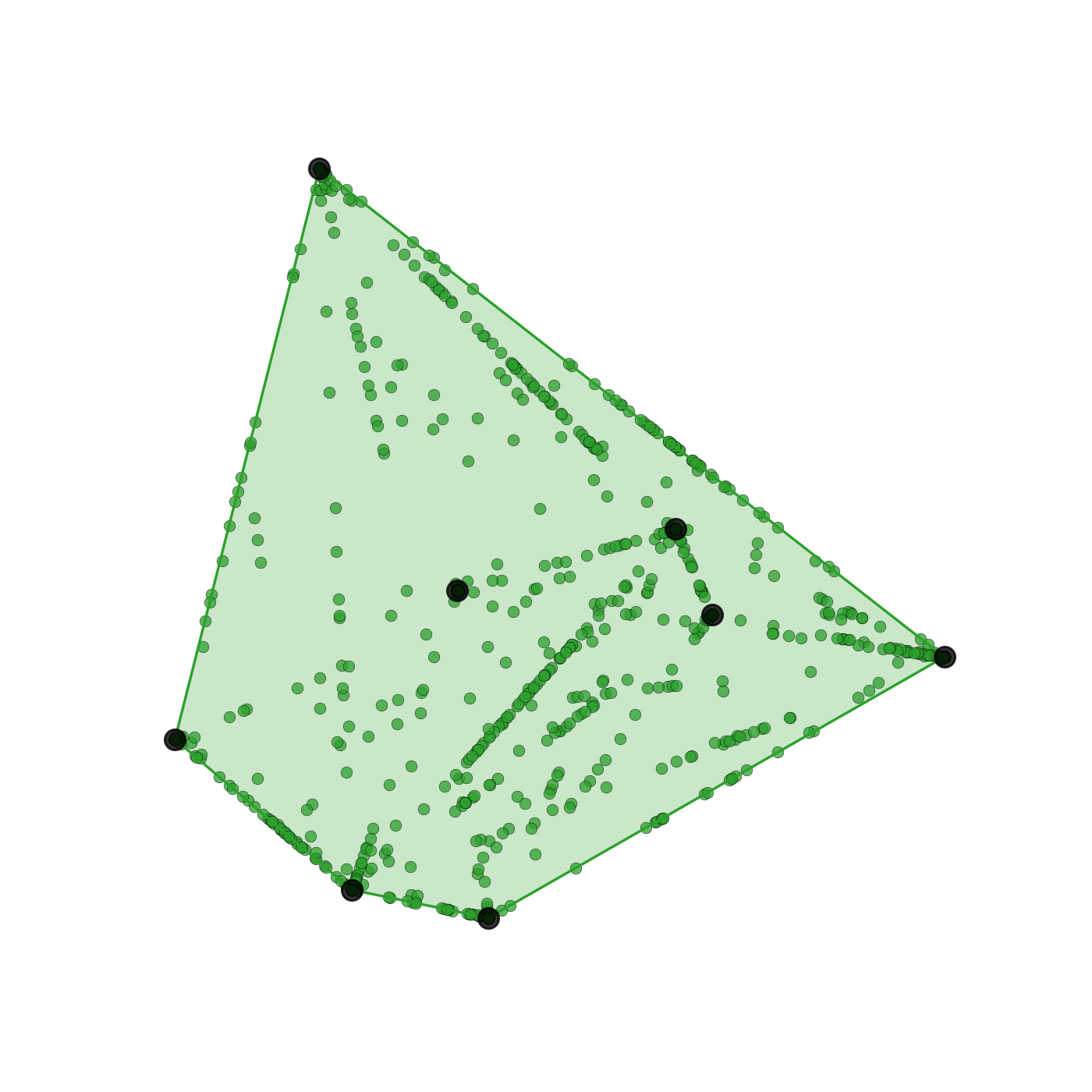}
    \caption{PCA: $\bm{B}_3$}
    \label{fig:local_hull_2}
\end{subfigure}
\hfill
\begin{subfigure}{0.11\textwidth}  
    \centering
    \includegraphics[width=1\linewidth]{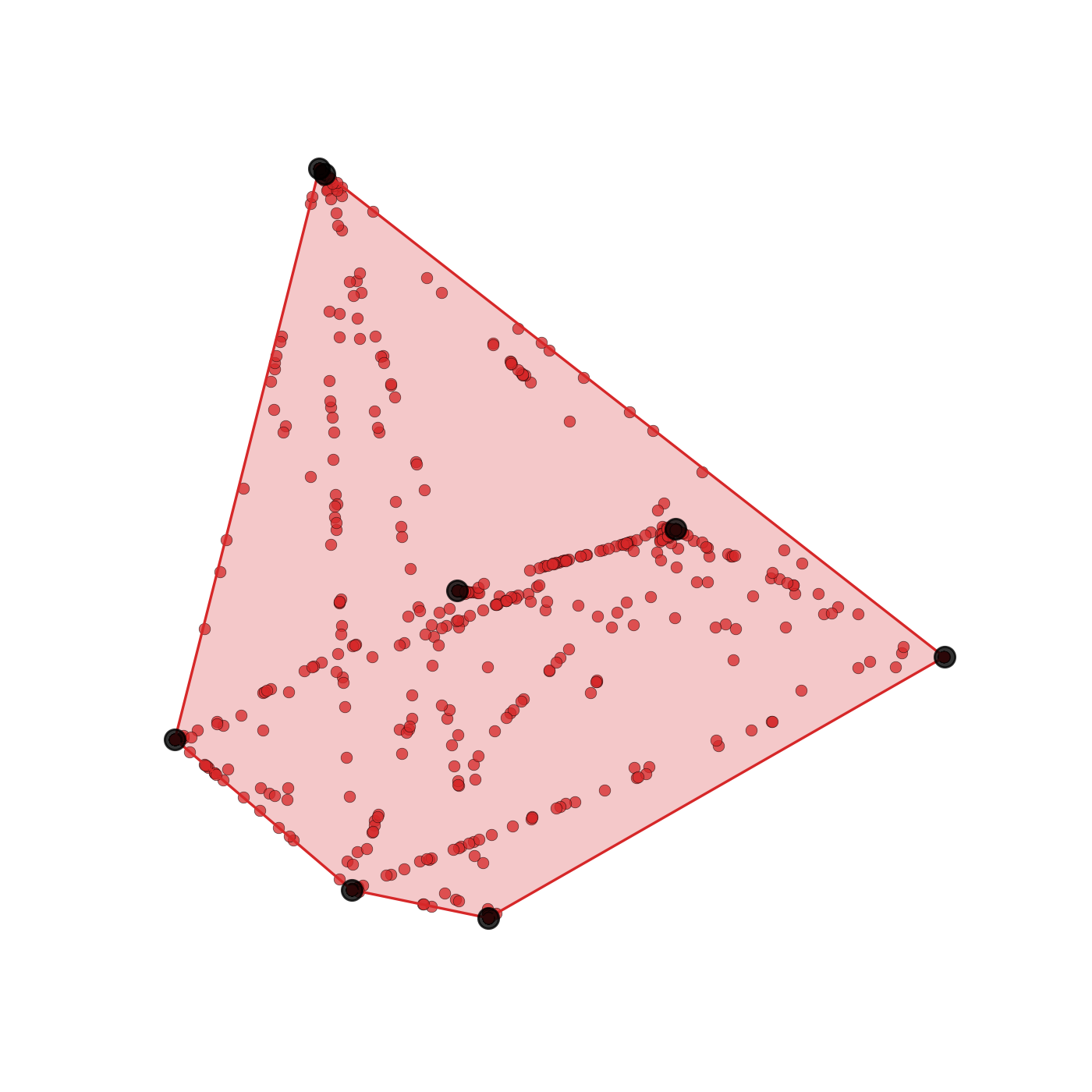}
    \caption{PCA: $\bm{B}_4$}
    \label{fig:local_hull_3}
\end{subfigure}
\hfill
\centering
\begin{subfigure}{0.11\textwidth}  
    \centering
    \includegraphics[width=1\linewidth]{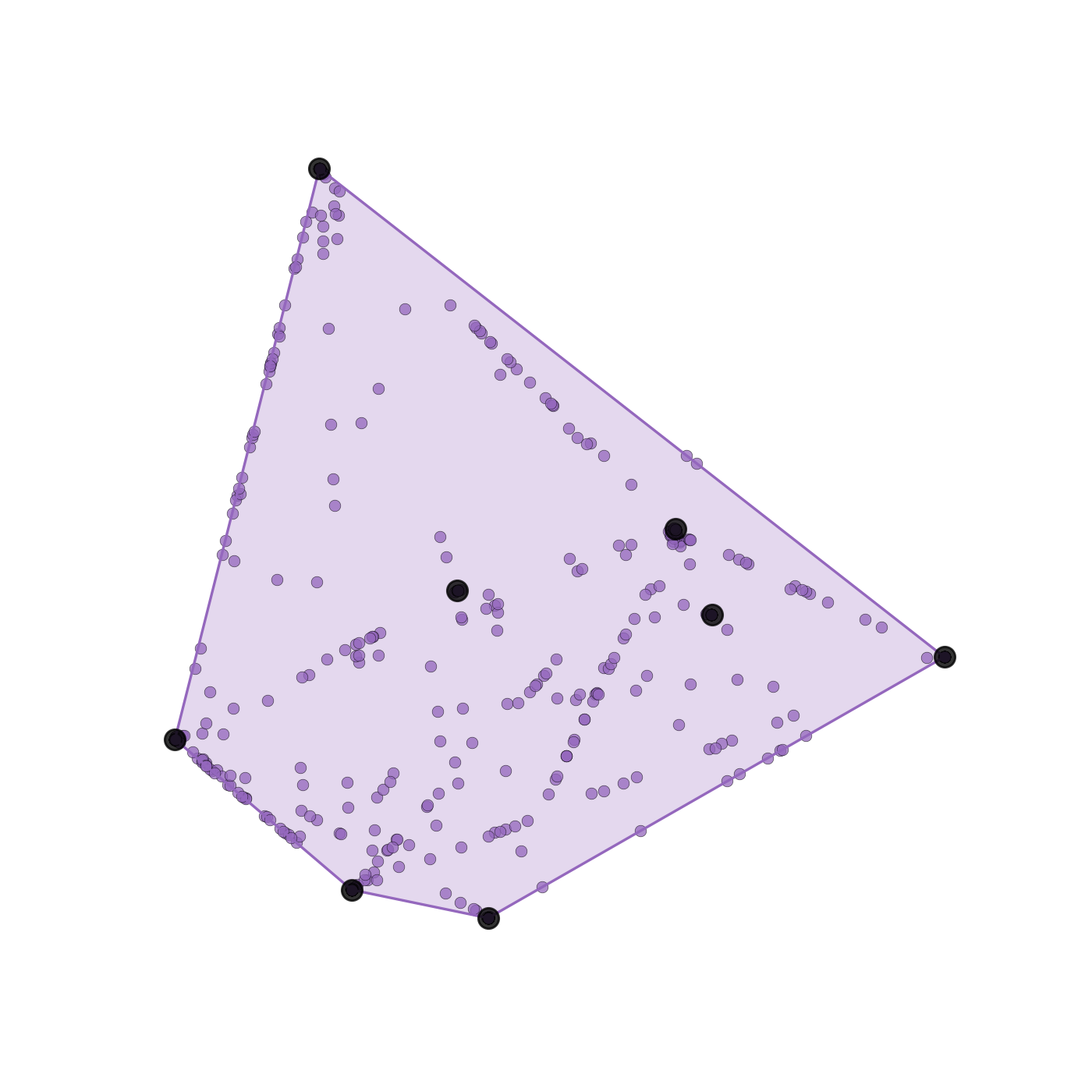}
    \caption{PCA: $\bm{B}_5$}
    \label{fig:local_hull_4}
\end{subfigure}
\hfill
\begin{subfigure}{0.11\textwidth}  
    \centering
    \includegraphics[width=1\linewidth]{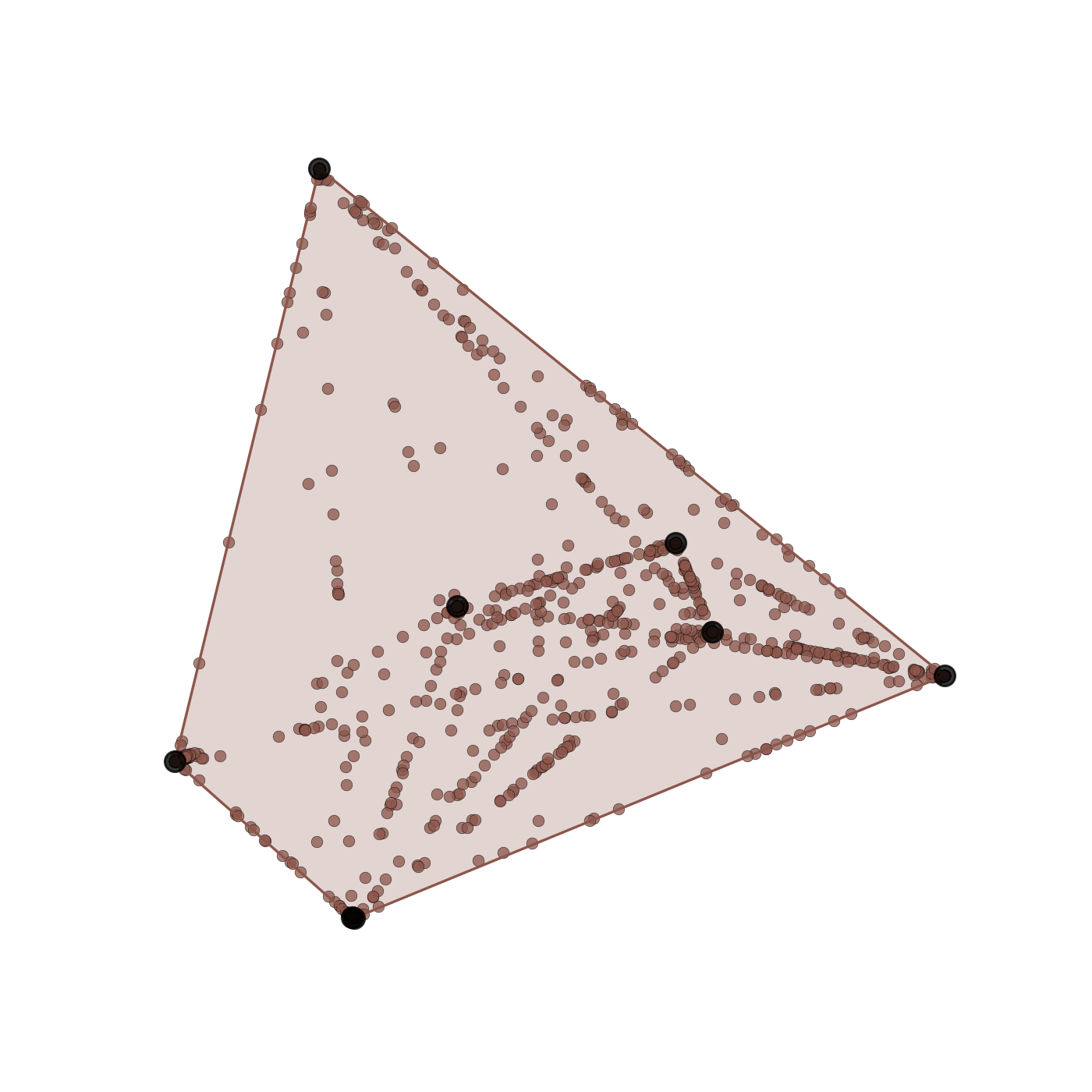}
    \caption{PCA: $\bm{B}_6$}
    \label{fig:local_hull_5}
\end{subfigure}
\hfill
\begin{subfigure}{0.11\textwidth}  
    \centering
    \includegraphics[width=1\linewidth]{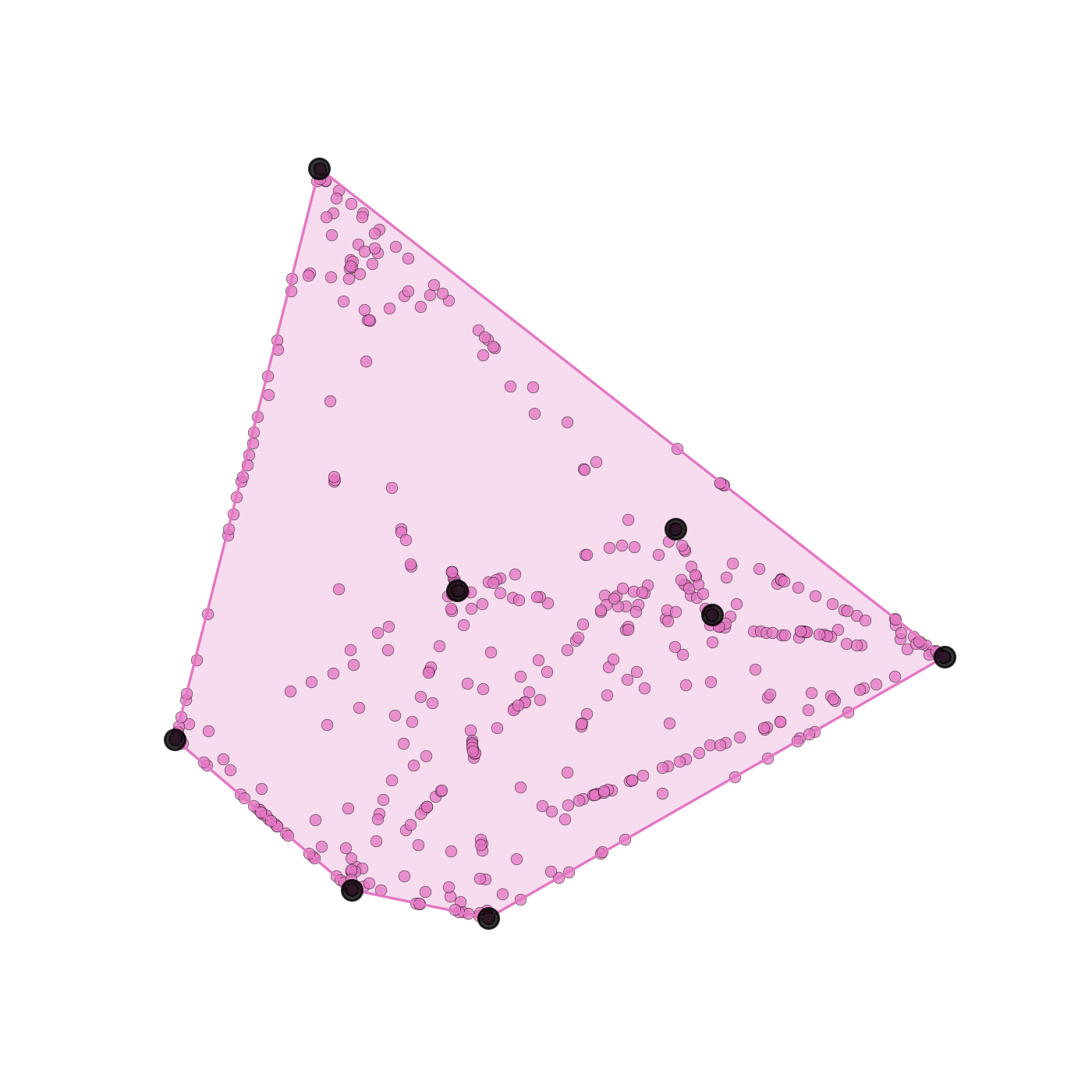}
    \caption{PCA: $\bm{B}_7$}
    \label{fig:local_hull_6}
\end{subfigure}
\hfill
\begin{subfigure}{0.11\textwidth}  
    \centering
    \includegraphics[width=1\linewidth]{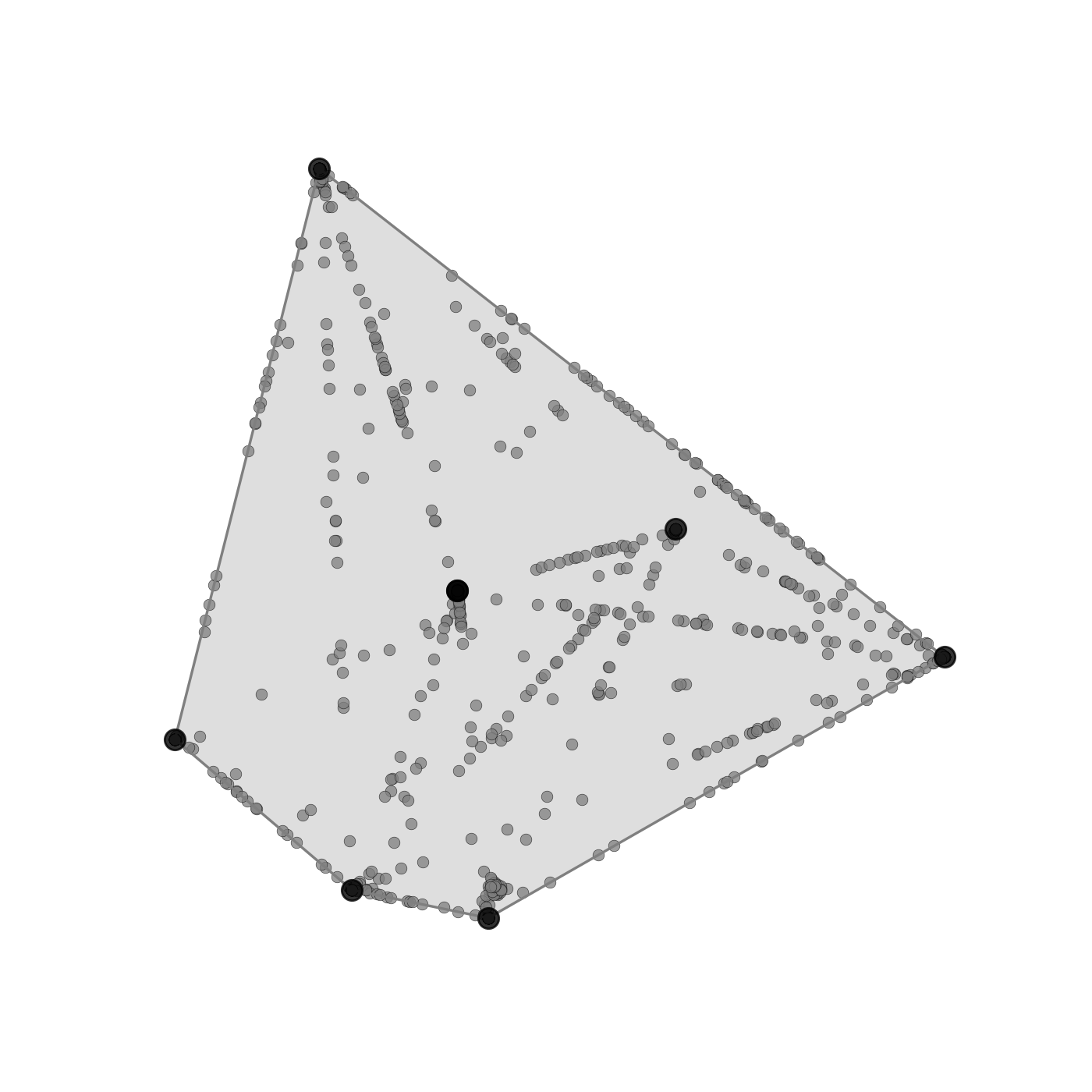}
    \caption{PCA: $\bm{B}_8$}
    \label{fig:local_hull_7}
\end{subfigure}

\hfill
\centering
\begin{subfigure}{0.115\textwidth}  
    \centering
    \includegraphics[width=1\linewidth]{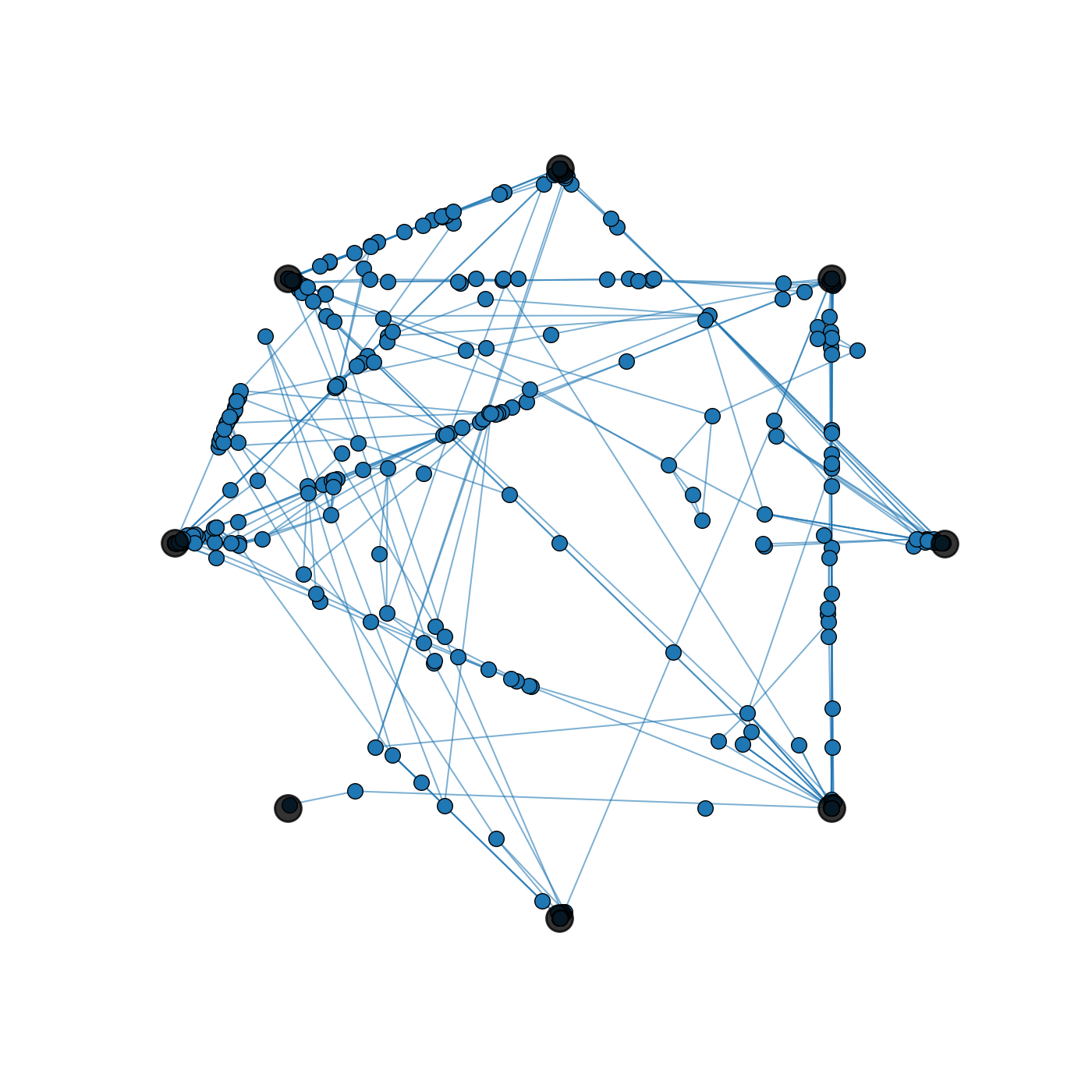}
    \caption{\textsc{Cp}: $\bm{B}_1$}
    \label{fig:b1}
\end{subfigure}
\hfill
\begin{subfigure}{0.115\textwidth}  
    \centering
    \includegraphics[width=1\linewidth]{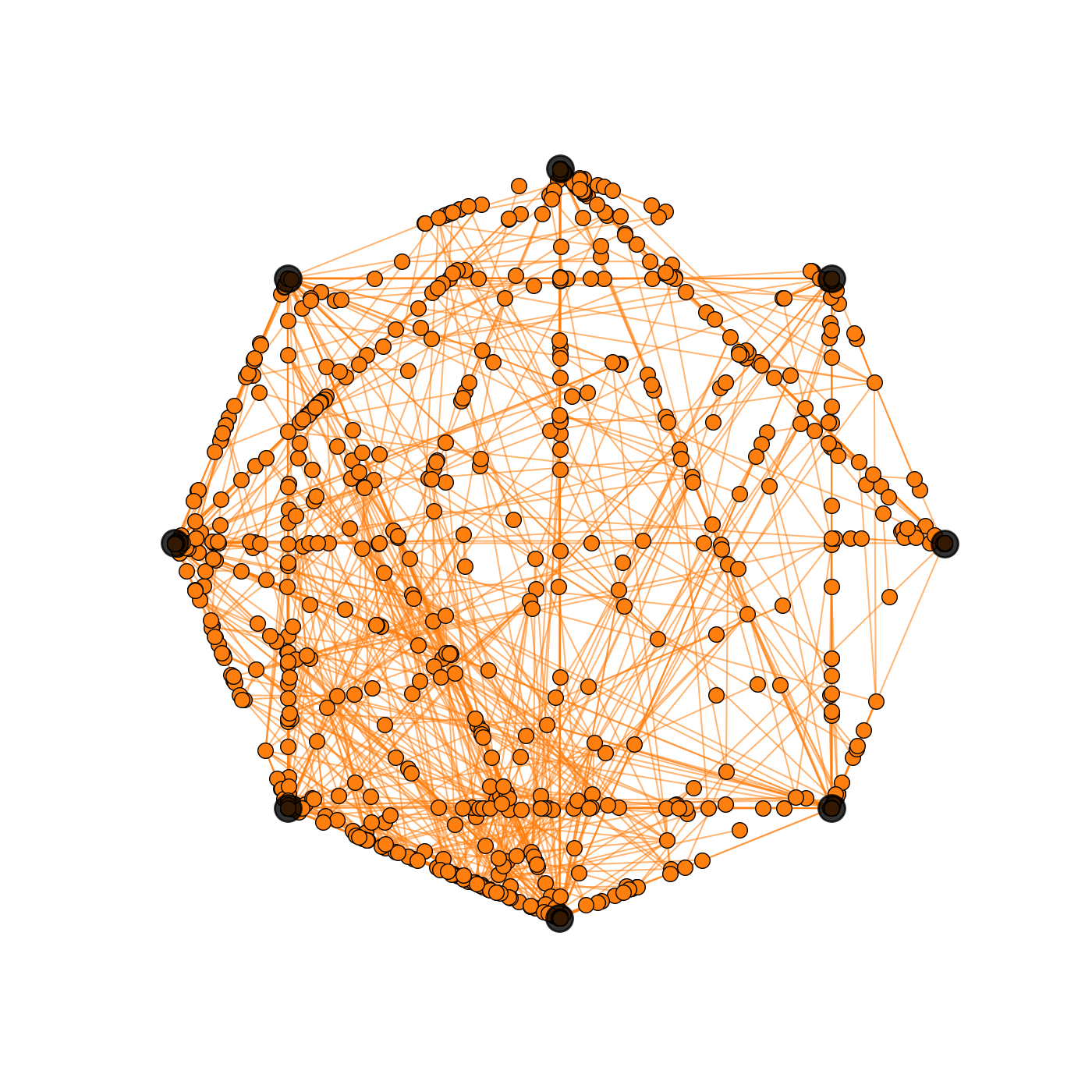}
    \caption{\textsc{Cp}: $\bm{B}_2$}
    \label{fig:b2}
\end{subfigure}
\hfill
\begin{subfigure}{0.115\textwidth}  
    \centering
    \includegraphics[width=1\linewidth]{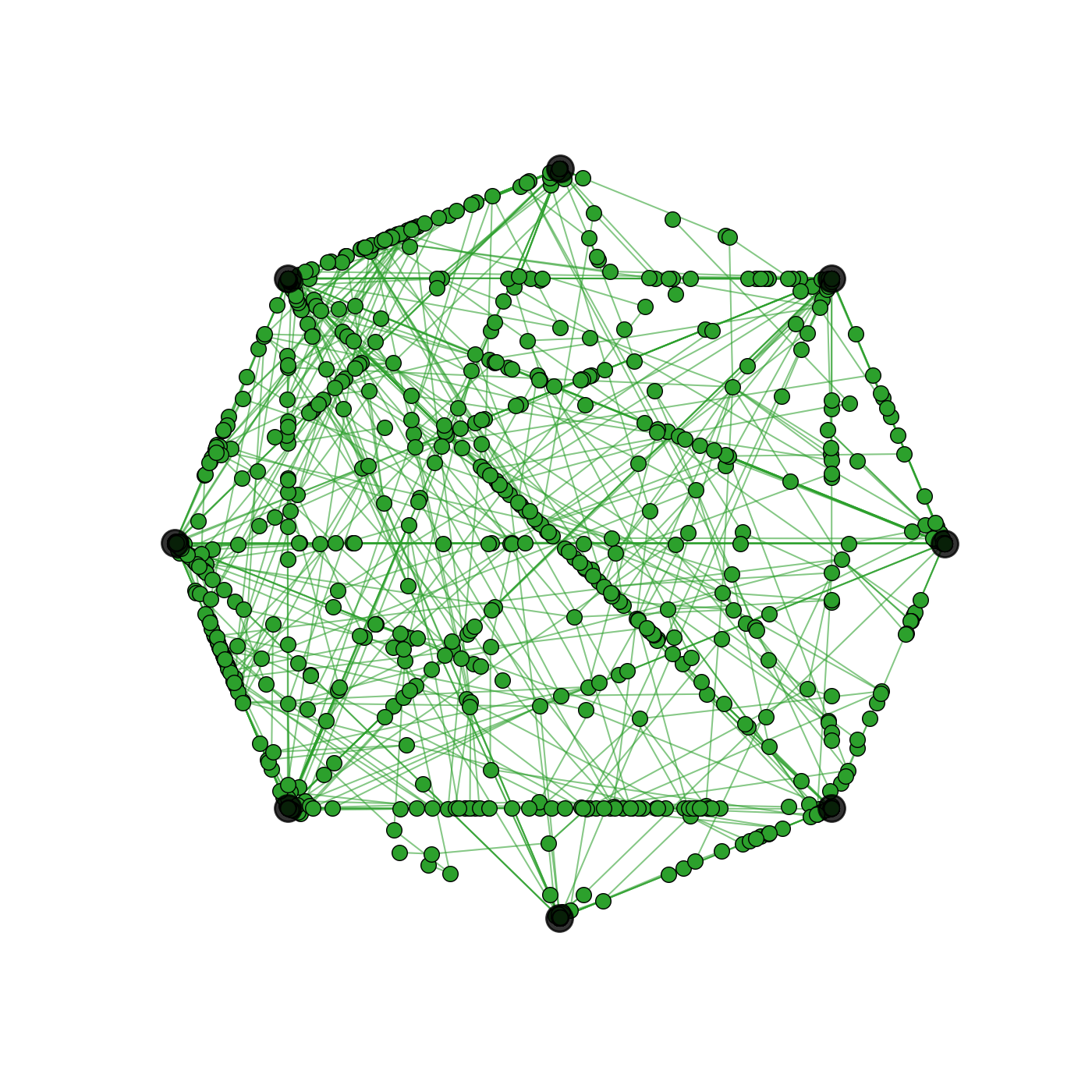}
    \caption{ \textsc{Cp}: $\bm{B}_3$}
    \label{fig:b3}
\end{subfigure}
\hfill
\begin{subfigure}{0.115\textwidth}  
    \centering
    \includegraphics[width=1\linewidth]{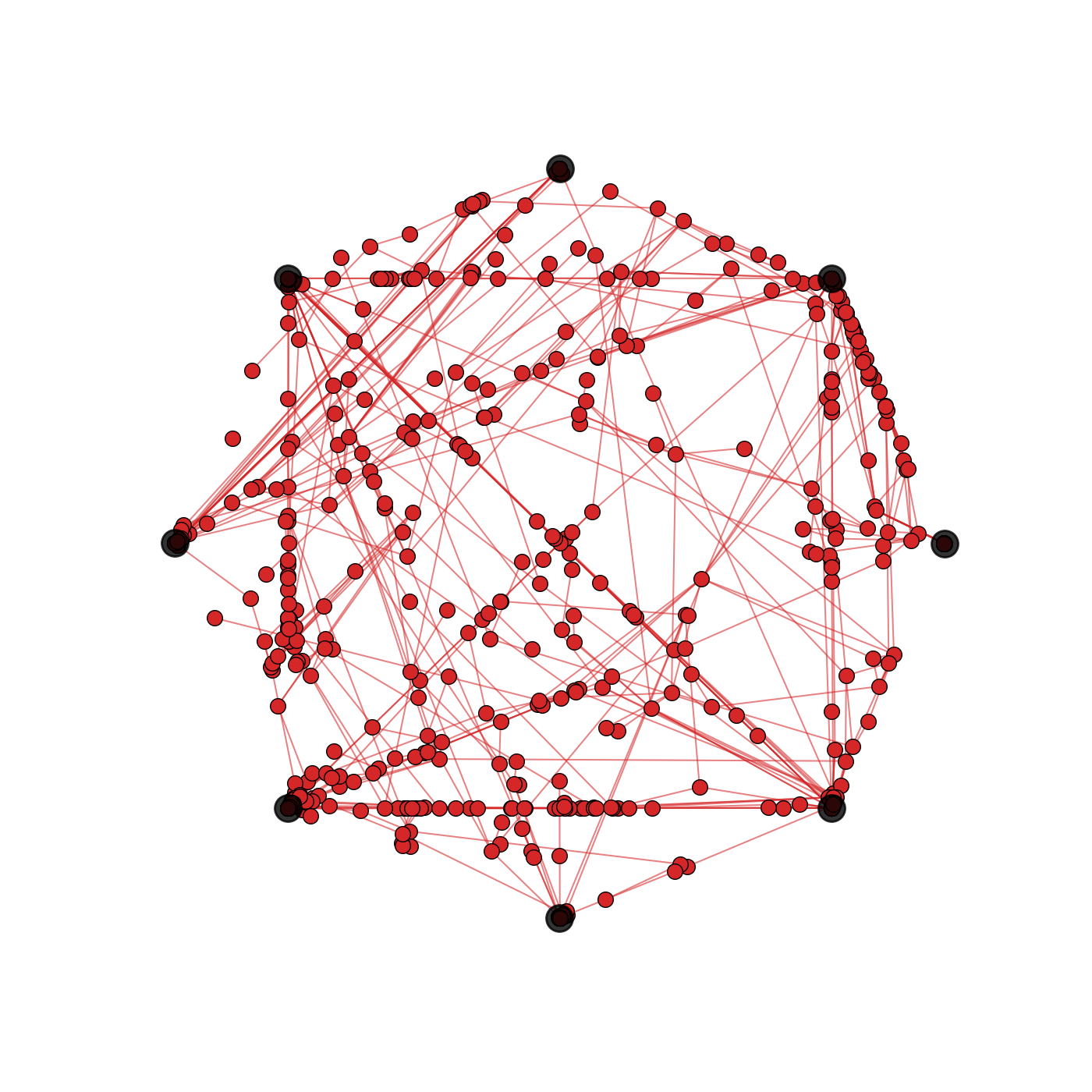}
    \caption{\textsc{Cp}: $\bm{B}_4$}
    \label{fig:b4}
\end{subfigure}
\hfill
\centering
\begin{subfigure}{0.115\textwidth}  
    \centering
    \includegraphics[width=1\linewidth]{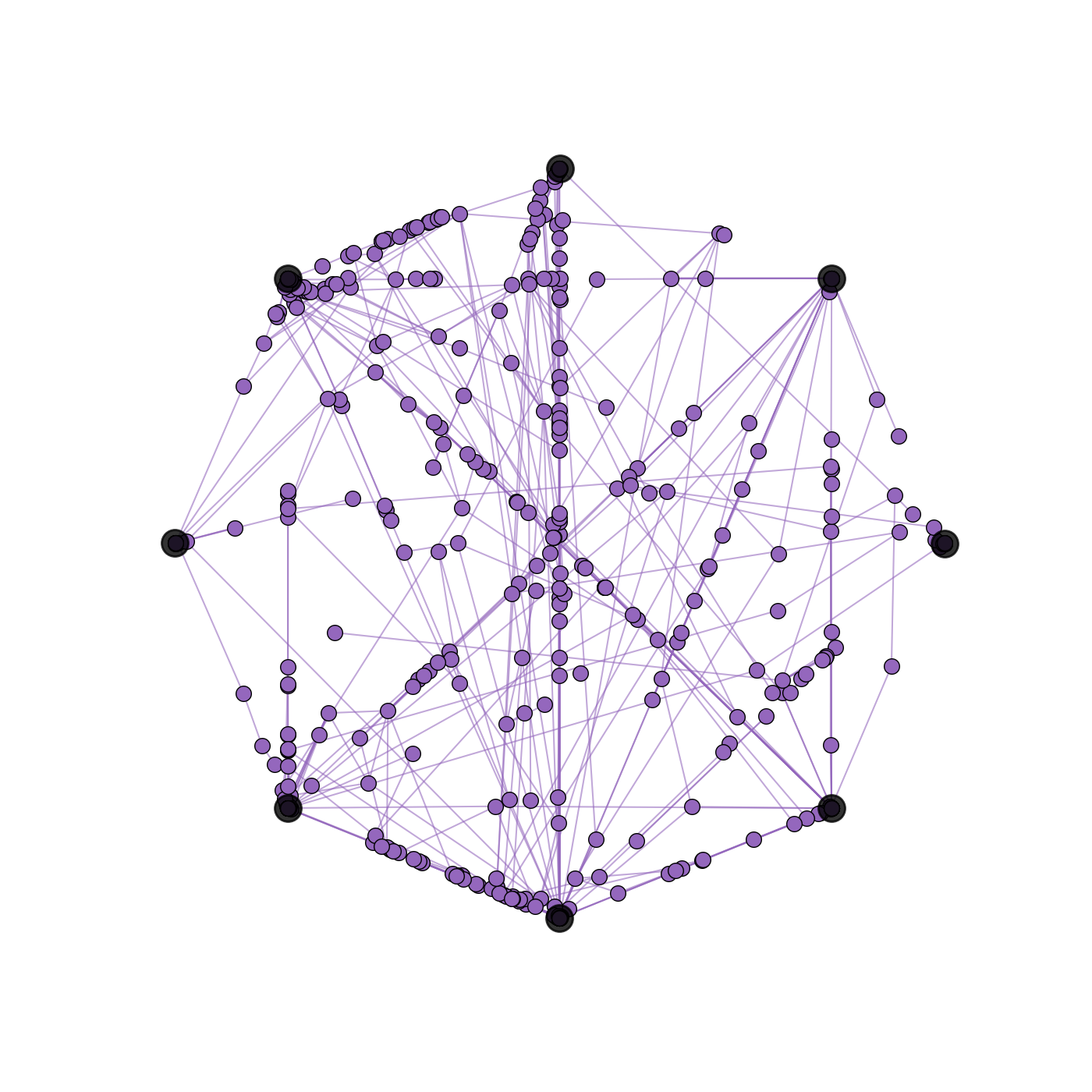}
    \caption{\textsc{Cp}: $\bm{B}_5$}
    \label{fig:b5}
\end{subfigure}
\hfill
\begin{subfigure}{0.115\textwidth}  
    \centering
    \includegraphics[width=1.\linewidth]{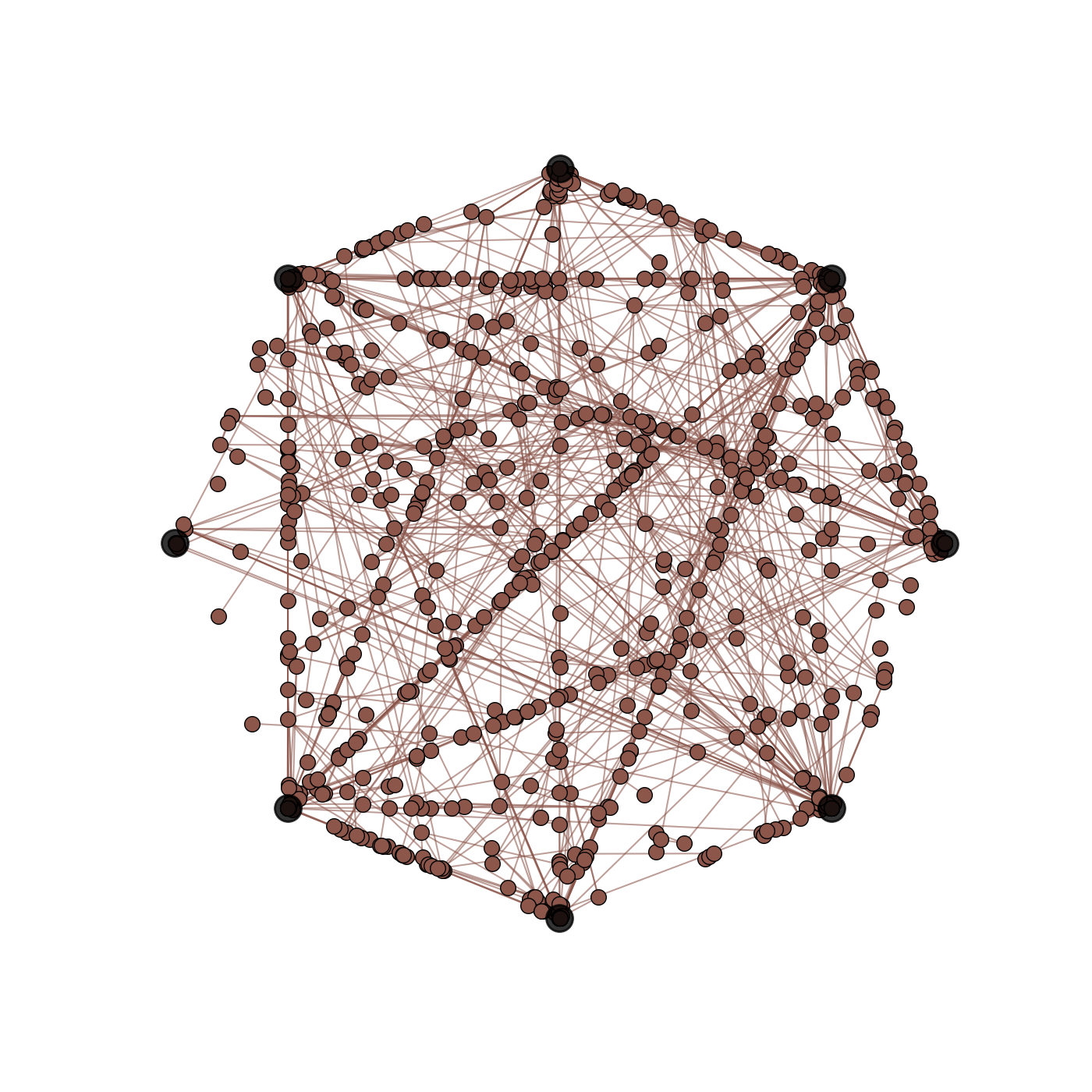}
    \caption{\textsc{Cp}: $\bm{B}_6$}
    \label{fig:b6}
\end{subfigure}
\hfill
\begin{subfigure}{0.115\textwidth}  
    \centering
    \includegraphics[width=1.\linewidth]{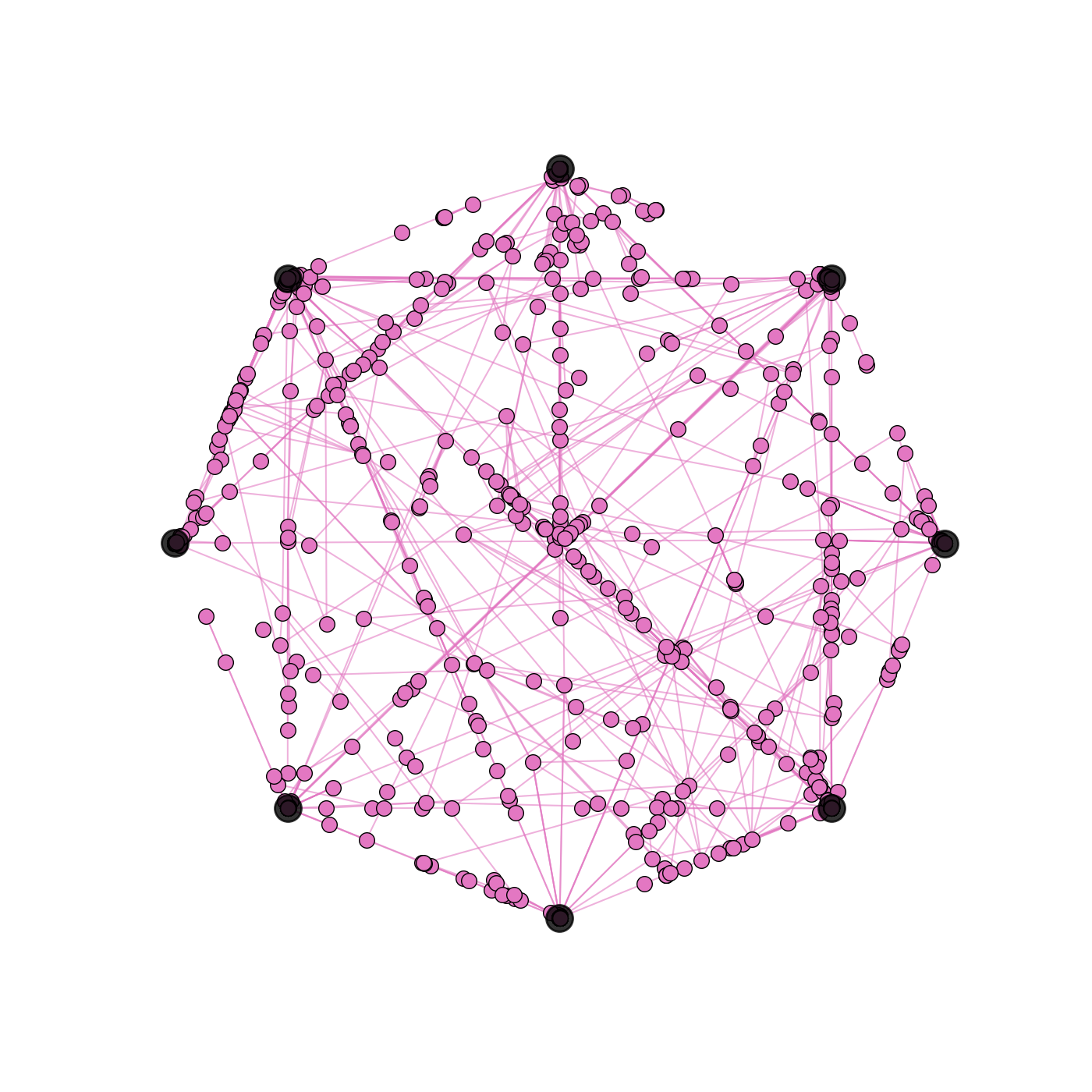}
    \caption{\textsc{Cp}: $\bm{B}_7$}
    \label{fig:b7}
\end{subfigure}
\hfill
\begin{subfigure}{0.115\textwidth}  
    \centering
    \includegraphics[width=1.\linewidth]{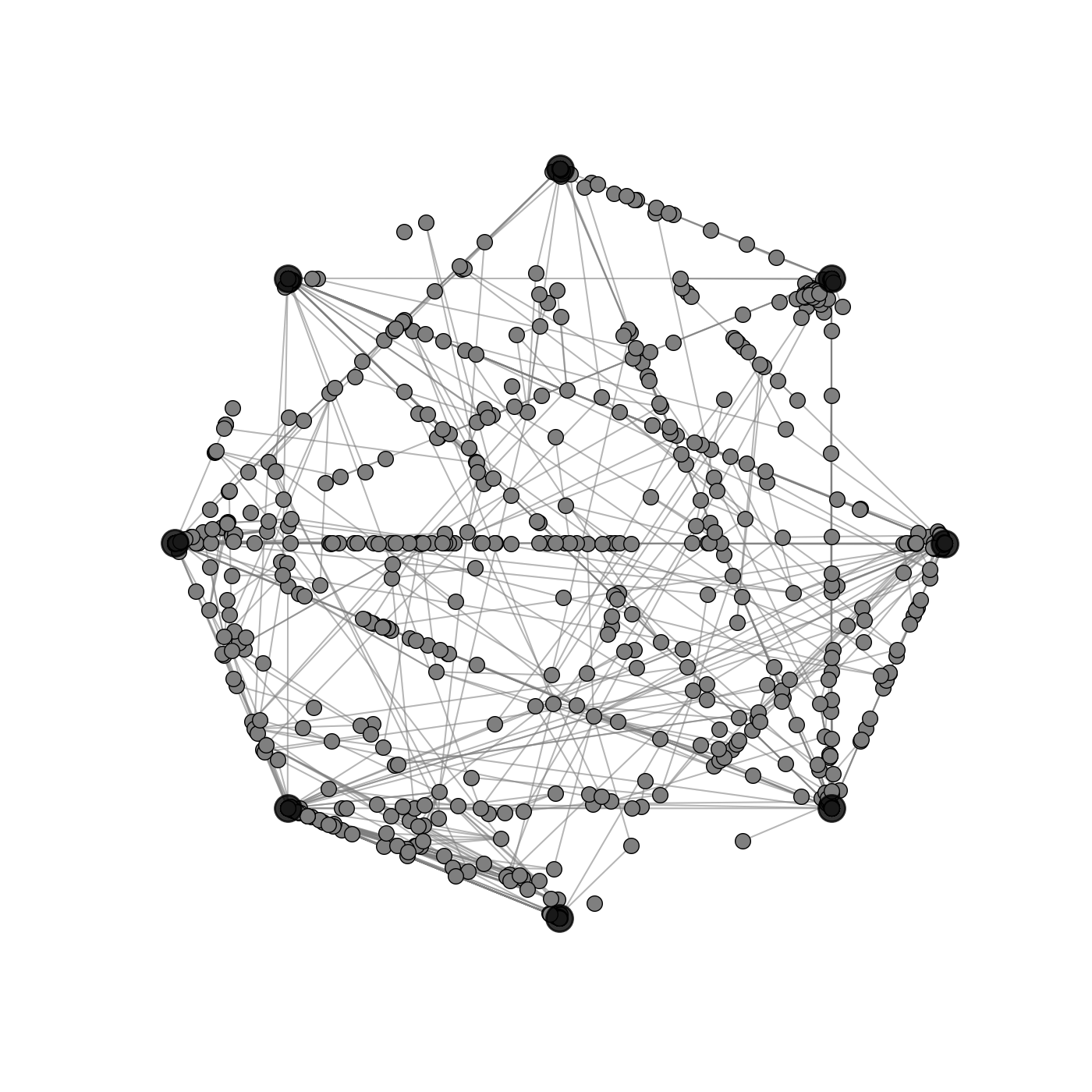}
    \caption{\textsc{Cp}: $\bm{B}_8$}
    \label{fig:b8}
\end{subfigure}
\caption{\textsc{GraphHull Visualizations:} 
(a) PCA projection of $K{=}8$ hulls in $D{=}8$ dimensions. 
(b) Reordered adjacency matrix: first by global hull assignments, then within blocks by $\arg\max \bm{\omega}_i$. 
(c)--(j) PCA projections of individual local hulls. 
(k)--(r) Prototype membership spaces for each $\bm{B}_k$, shown as circular plots (\textsc{Cp}) with prototypes evenly spaced on the circle, with lines indicating links inside a hull (colors denote local hulls).}
\label{fig:hepth_visualization}
\end{figure*}

\textbf{Link prediction.}
For link prediction, we follow the widely adopted evaluation protocol of \citet{perozzi2014deepwalk,nakis2023hierarchicalblockdistancemodel}. 
Specifically, we randomly remove $50\%$ of the edges while ensuring that the residual graph remains connected. The removed edges, together with an equal number of randomly sampled non-edges, form the positive and negative instances of the test set. The residual graph is then used to learn node embeddings. We evaluate performance on five benchmark networks, each over five runs and across multiple embedding dimensions ($D \in \{8,16,32,64\}$). Table~\ref{tab:AUC} reports the Area Under the Receiver Operating Characteristic Curve (AUC-ROC) (for Precision-Recall (PR-AUC) scores see supplementary). Across runs, the variance was consistently on the order of $10^{-3}$ and is omitted for readability. Following \citet{grover2016node2vec}, dyadic features are constructed using binary operators (average, Hadamard, weighted-$L_1$, weighted-$L_2$) and a logistic regression classifier with $L_2$ regularization is trained to make predictions. In contrast, predictions from \textsc{GraphHull} $(\epsilon=0.49)$ are obtained directly from the model: we use the log-odds $\eta_{ij}$ of a test pair $\{i,j\}$ to compute link probabilities, with no additional classifier required. Importantly, \textsc{GraphHull} is tuned solely on training sets, and remains blind to the test set. By comparison, many baseline results are reported under test-aware tuning. Despite this advantage, \textsc{GraphHull} remains highly competitive with strong baselines across datasets and embedding dimensions, achieving comparable or superior performance in many settings while maintaining consistent link prediction accuracy.

\textbf{Community detection.} To assess community recovery, we use four networks with ground-truth labels. 
For membership-aware models, including \textsc{GraphHull}, we set the latent dimension equal to the number of true communities and compare inferred memberships with the ground truth. For GRL methods without memberships, we extract embeddings of the same dimensionality and apply $k$-means. We report Normalized Mutual Information (NMI) and Adjusted Rand Index (ARI) averaging over five runs. All baselines are tuned for their main hyperparameters, while \textsc{GraphHull} uses the same settings across datasets. Results in Table~\ref{tab:nmi_ari} show \textsc{GraphHull} is consistently competitive with GRL baselines and outperforms membership-based methods, with the exception of \textsl{Cora}, where GRL with post-hoc clustering has an advantage. Nevertheless, \textsc{GraphHull} is the most consistent across datasets and metrics, outperforming membership-based methods, remaining competitive with GRL approaches.

 \textbf{Network visualization.} We illustrate how \textsc{GraphHull} captures global archetypal and local prototypical structures in the \textsl{HepTh} network. In Figure~\ref{fig:hepth_visualization}, panel (a) shows the PCA projection of the latent space inferred by \textsc{GraphHull}. Each local hull $\bm{B}_k$ is anchored in a distinct global archetype defined by the global hull $\bm{A}$. Any apparent overlap in panel (a) arises only from projecting to two dimensions; in the full $D{=}8$ space, all hulls are guaranteed to be non-overlapping. Panel (b) shows the adjacency matrix reordered by global hull assignments, and within each block by the maximum prototype membership $\argmax \bm{\omega}_i$. This reveals clear block structure at both the global and local levels. Panels (c)--(j) present disentangled PCA projections of each local hull $\bm{B}_k$. Finally, panels (k)--(r) show circular plots of the prototype membership spaces, where prototypes are positioned every $\tfrac{2\pi}{K}$ radians, illustrating node memberships $\bm{\omega}_i$ enriched with the links across nodes inside each block. Overall, \textsc{GraphHull} extracts informative and robust network structures while combining the strengths of global archetypal characterization with prototypical structure inside communities.
\begin{table*}[!t]
\centering
\caption{Normalized Mutual Information (NMI) and Adjusted Rand Index (ARI) scores. 
}
\label{tab:nmi_ari}
\resizebox{0.85\textwidth}{!}{%
\begin{tabular}{lcccccccc}\toprule
\multicolumn{1}{l}{} & \multicolumn{2}{c}{\textsl{Cora}} & \multicolumn{2}{c}{\textsl{Citeseer}} & \multicolumn{2}{c}{\textsl{LastFM}}& \multicolumn{2}{c}{\textsl{Pol}}\\\cmidrule(rl){2-3}\cmidrule(rl){4-5}\cmidrule(rl){6-7}\cmidrule(rl){8-9}
\multicolumn{1}{c}{Metric} & NMI & ARI & NMI & ARI & NMI & ARI & NMI & ARI
\\\cmidrule(rl){1-1}\cmidrule(rl){2-2}\cmidrule(rl){3-3}\cmidrule(rl){4-4}\cmidrule(rl){5-5}\cmidrule(rl){6-6}\cmidrule(rl){7-7}\cmidrule(rl){8-8}\cmidrule(rl){9-9}
\textsc{Node2Vec}&\textbf{.460} $\pm$ .010	& \textbf{.397} $\pm$ .024& .223 $\pm$ .010  &.211 $\pm$ .025 	&.591 $\pm$ .005	& .436 $\pm$ .009  & \uline{.727} $\pm$ .007  &  \uline{.791} $\pm$ .008  \\
\textsc{Role2Vec}  &\uline{.486} $\pm$ .001	&\uline{.417} $\pm$ .010 & .299 $\pm$ .016  & .279 $\pm$ .015 	&.596 $\pm$ .001	& .455 $\pm$ .003  &  .712 $\pm$ .003 &.772 $\pm$ .003 \\
\textsc{NetMF} &.463 $\pm$ .006	& .349 $\pm$ .006 & .223 $\pm$ .010   & .145 $\pm$ .016	&\uline{.605} $\pm$ .001	& \textbf{.522} $\pm$ .001  & .145 $\pm$ .001  &.199 $\pm$ .001 \\
\textsc{GraRep} &.292 $\pm$ .001& .192 $\pm$ .001 &\uline{.320} $\pm$ .001   &\uline{.320} $\pm$ .001 	&.394 $\pm$ .001	& .259 $\pm$ .001  & .447 $\pm$ .001  & .492 $\pm$ .001 \\
\textsc{RandNE}&.020 $\pm$ .004	&.010 $\pm$ .004  & .015 $\pm$ .002  & .010 $\pm$ .003	&.063 $\pm$ .001	& .018 $\pm$ .001  & .003 $\pm$ .001  &.003 $\pm$ .001 \\\midrule
\textsc{NNSED}  & .310 $\pm$ .031  & .229 $\pm$ .027 &.244 $\pm$ .011	& .201 $\pm$ .016 	&.290 $\pm$ .006	& .143 $\pm$ .011  & .085 $\pm$ .043  &.136 $\pm$ .056   \\
\textsc{MNMF}   &.294 $\pm$ .017	&.226 $\pm$ .018 & .144 $\pm$ .031  &.125 $\pm$ .030 	&.473 $\pm$ .006	& .307 $\pm$ .012  & .599 $\pm$ .024  & .624 $\pm$ .030  \\
\textsc{SymmNMF}  &	.311 $\pm$ .001&.204 $\pm$ .001 & .310 $\pm$ .006  &.289 $\pm$ .003 	&.448 $\pm$ .020	&.329 $\pm$ .043   & .117 $\pm$ .006  & .153 $\pm$ .004   \\ \bottomrule
\textsc{Dmon}  &.350 $\pm$	.024&.299 $\pm$ .042&.271 $\pm$ 0.018   &.265 $\pm$ 0.027	&.518 $\pm$	.025&.397  $\pm$ .049&.714 $\pm$  .001&.769 $\pm$ .001\\
\midrule
\textsc{GraphHull} &.422 $\pm$ .004	&.333 $\pm$ .008 &\textbf{.347} $\pm$ .007   &\textbf{.372} $\pm$ .009 	&\textbf{.616} $\pm$ .001	& \uline{.514} $\pm$ .003  & \textbf{.757} $\pm$ .001 & \textbf{.820} $\pm$ .001
\\
\bottomrule    
\end{tabular}%
}
\end{table*}
 \begin{figure}[!t]
\centering
\begin{subfigure}{0.22\textwidth}  
    \centering
    \includegraphics[width=1\linewidth]{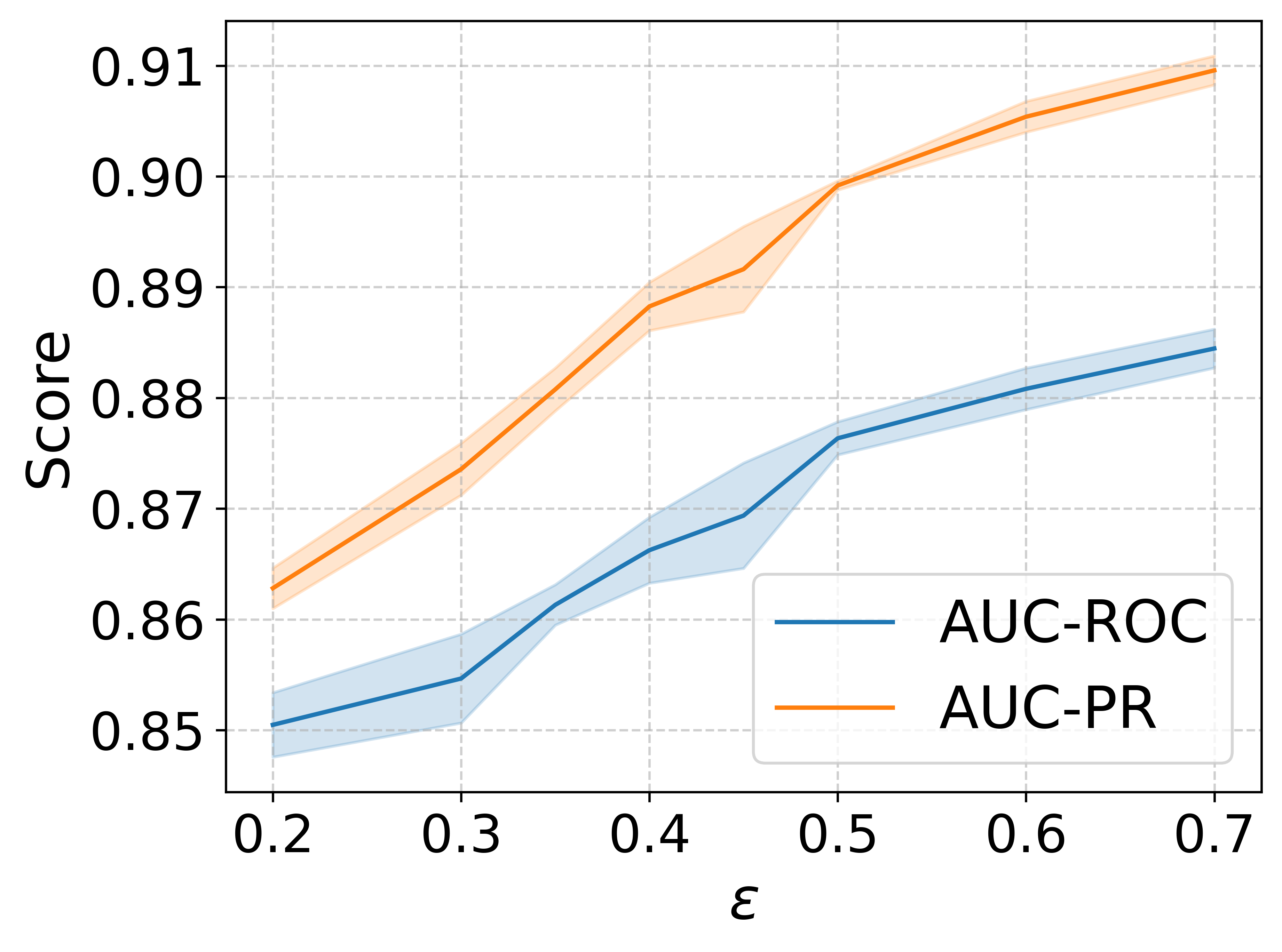}
    \caption{Predictive performance as a function of $\varepsilon$.}
    \label{fig:AUC_epsilon}
\end{subfigure}
\hfill
\begin{subfigure}{0.245\textwidth}  
    \centering
    \includegraphics[width=1\linewidth]{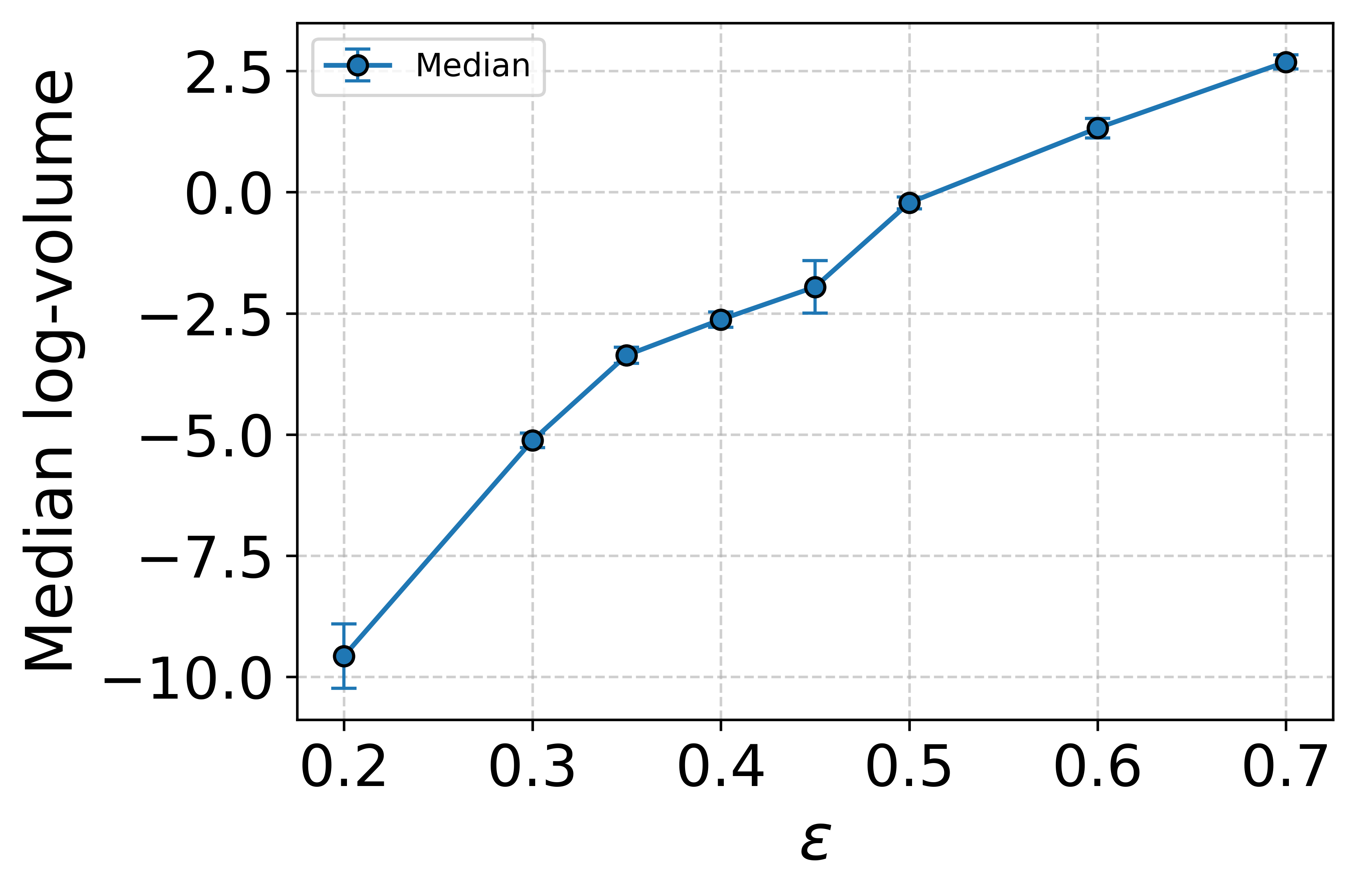}
    \caption{Effective log-volume of hulls as a function of $\varepsilon$.}
    \label{fig:vols_epsilon}
\end{subfigure}
\begin{subfigure}{0.46\textwidth}  
    \centering
    \includegraphics[width=1\linewidth]{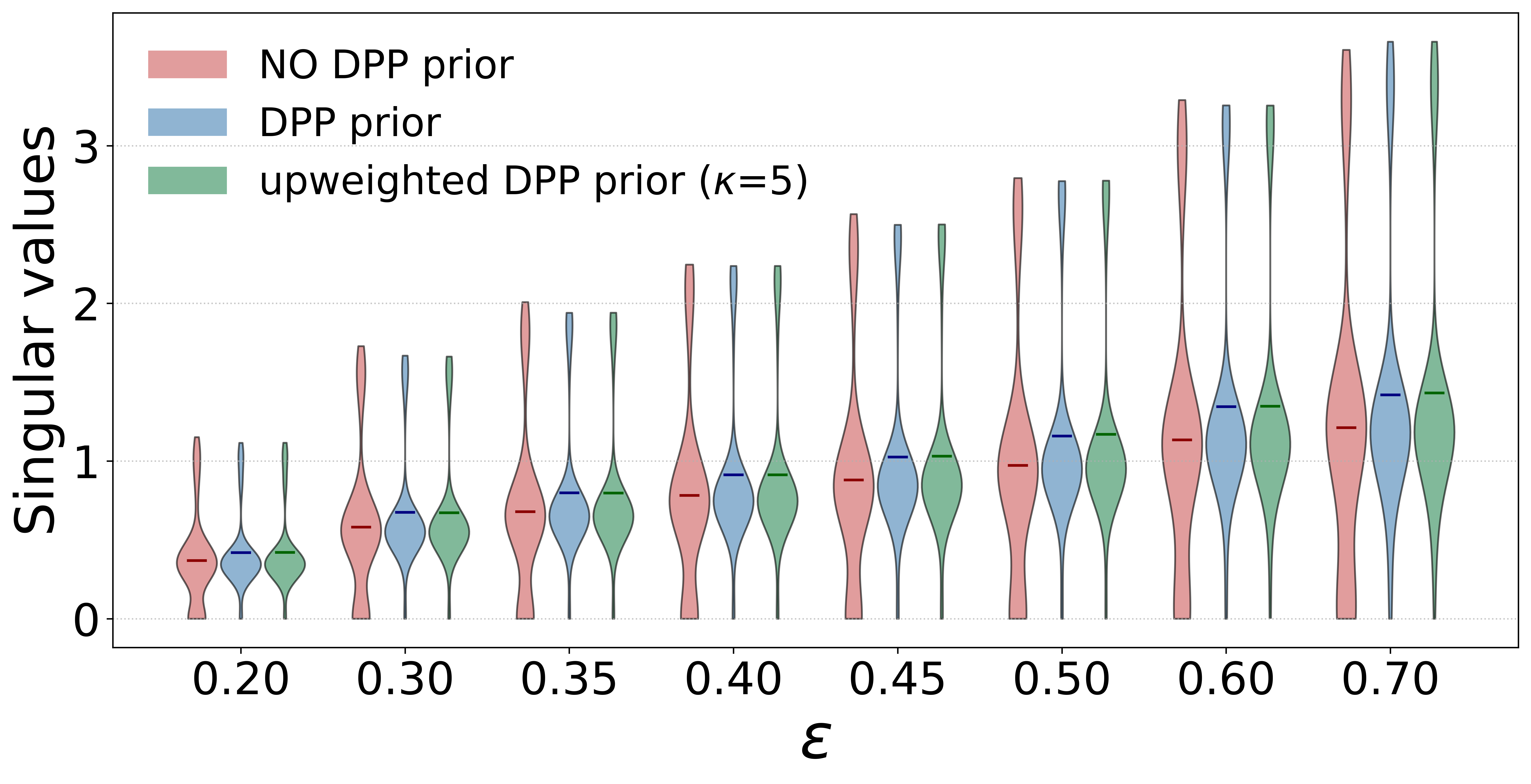}
    \caption{Violin plots of local-hull singular values as a function of the anchor strength $\varepsilon$ under three priors: none (no DPP), standard DPP, and upweighted DPP ($\kappa=5$).}
    \label{fig:dpp_epsilon}
\end{subfigure}
\caption{\textsc{Effect of $\varepsilon$:} Results on the \textsl{HepTh} dataset.}
\label{fig:2pols}
\end{figure}

\textbf{Effect of anchor strength $\bm{\varepsilon}$.}
The parameter $\varepsilon$ controls the maximum spread of each hull; for $\varepsilon<\tfrac{1}{2}$ it guarantees disjointness. Figure~\ref{fig:AUC_epsilon} shows link prediction performance as a function of $\varepsilon$. As $\varepsilon$ increases, predictive performance steadily improves in both AUC-ROC and AUC-PR. This trend continues even in the overlapping regime ($\varepsilon>\tfrac{1}{2}$), where allowing overlaps better explains mixed-memberships. Notably, at $\varepsilon \approx 0.45$ we achieve high predictive accuracy and remain within the identifiable regime. We investigate hull geometry as $\varepsilon$ varies. Figure~\ref{fig:vols_epsilon} reports the effective log-volume across runs: volumes increase monotonically with $\varepsilon$, showing that hulls naturally expand as anchor dominance relaxes. Although the DPP prior discourages coincident prototypes, it does not enforce full affine independence per hull. In practice, some communities form hulls of \emph{reduced effective rank}, where a prototype is nearly an affine combination of others. This reflects benign over-parameterization: the data does not require all directions. Such rank deficiencies are harmless, the redundant vertex can be trivially removed without changing the hull, providing an interpretable estimate of a community's intrinsic dimension.
Concretely, replacing $\bm{L}_k$ with $\kappa \bm{L}_k$ for $\kappa>0$ scales the eigenvalues of the kernel, yielding the prior $\log\det(\kappa\bm{L}_k)-\log\det(\bm{I}+\kappa\bm{L}_k)$. Larger $\kappa$ strengthens the repulsion between prototypes, discouraging redundant vertices without altering the feasible set of hulls. This effect is evident in Figure~\ref{fig:dpp_epsilon}: without a DPP prior, the singular values of local hulls $\bm{B}_k$ exhibit heavy tails near zero, producing unstable hulls with rank deficiency. Including a DPP prior almost entirely eliminates this issue, while upweighting with $\kappa=5$ further removes rank deficiency.

\textbf{Prototype interpretability.}
Examining the learned prototypes reveals clear structural roles. Across datasets, anchors generally correspond to dense cores, while non-anchor prototypes capture sparser or peripheral structures, highlighting diversity within each community. For example, in the \textsl{HepTh} network anchors exhibit systematically higher clustering coefficients than non-anchor prototypes, whereas peripheral prototypes often attract large numbers of ``on-vertex'' exemplars despite low clustering values. These findings confirm that the anchor–prototype hierarchy yields interpretable profiles of structural organization (see supplementary for detailed quantitative analysis).

\section{Conclusion \& Limitations}
We introduced \textsc{GraphHull}, a principled generative framework for explainable graph representation learning. By combining global archetypes with anchor–dominant local convex hulls, our approach yields embeddings that are identifiable, interpretable, and diverse by design. The geometry enforces disjointness and stability, while determinantal
point process priors promote non-degenerate structure. Across multiple networks, GraphHull achieves competitive or superior performance on link prediction and community detection, while naturally providing multi-scale explanations of communities and prototypes. Like most generative models, optimization is non-convex and depends on initialization, but in practice we find \textsc{GraphHull} to be stable under deterministic initializations based on the normalized Laplacian. Beyond graph analysis, the proposed hierarchical archetypal design is broadly applicable to other domains where transparent latent geometry is needed. We hope this work encourages further research into generative, geometry-aware models for interpretable machine learning.

\section*{Acknowledgements}
We gratefully acknowledge the reviewers for their constructive feedback and insightful comments. C. K. is supported by the IdAML Chair hosted at ENS Paris-Saclay, Université Paris-Saclay.

\bibliography{reference}

\section*{Checklist}

\begin{enumerate}

  \item For all models and algorithms presented, check if you include:
  \begin{enumerate}
    \item A clear description of the mathematical setting, assumptions, algorithm, and/or model. Yes 
    \item An analysis of the properties and complexity (time, space, sample size) of any algorithm. Yes
    \item (Optional) Anonymized source code, with specification of all dependencies, including external libraries. Yes as a supplementary file
  \end{enumerate}

  \item For any theoretical claim, check if you include:
  \begin{enumerate}
    \item Statements of the full set of assumptions of all theoretical results. Yes
    \item Complete proofs of all theoretical results. Yes in the supplementary
    \item Clear explanations of any assumptions. Yes
  \end{enumerate}

  \item For all figures and tables that present empirical results, check if you include:
  \begin{enumerate}
    \item The code, data, and instructions needed to reproduce the main experimental results (either in the supplemental material or as a URL). Yes in the supplementary
    \item All the training details (e.g., data splits, hyperparameters, how they were chosen). Yes
    \item A clear definition of the specific measure or statistics and error bars (e.g., with respect to the random seed after running experiments multiple times). Yes
    \item A description of the computing infrastructure used. (e.g., type of GPUs, internal cluster, or cloud provider). Yes
  \end{enumerate}

  \item If you are using existing assets (e.g., code, data, models) or curating/releasing new assets, check if you include:
  \begin{enumerate}
    \item Citations of the creator If your work uses existing assets. Yes
    \item The license information of the assets, if applicable. Not Applicable
    \item New assets either in the supplemental material or as a URL, if applicable. Not Applicable
    \item Information about consent from data providers/curators. Not Applicable
    \item Discussion of sensible content if applicable, e.g., personally identifiable information or offensive content. Not Applicable
  \end{enumerate}

  \item If you used crowdsourcing or conducted research with human subjects, check if you include:
  \begin{enumerate}
    \item The full text of instructions given to participants and screenshots. Not Applicable
    \item Descriptions of potential participant risks, with links to Institutional Review Board (IRB) approvals if applicable. Not Applicable
    \item The estimated hourly wage paid to participants and the total amount spent on participant compensation. Not Applicable
  \end{enumerate}

\end{enumerate}

\end{document}